%% file: 0_main.tex
\documentclass[letterpaper]{article} 
\usepackage[preprint]{aaai2027}  
\usepackage[hyphens]{url}  
\usepackage{graphicx} 
\usepackage{natbib}  
\usepackage{caption} 
\usepackage{algorithm}
\usepackage{algorithmic}

\usepackage{newfloat}
\usepackage{listings}
\DeclareCaptionStyle{ruled}{labelfont=normalfont,labelsep=colon,strut=off} 
\floatstyle{ruled}
\newfloat{listing}{tb}{lst}{}
\floatname{listing}{Listing}

\usepackage{booktabs}

\usepackage{bm}
\usepackage{amsmath}
\usepackage{amsfonts}
\usepackage{cleveref}
\usepackage{multirow}
\usepackage{subcaption}
\usepackage{booktabs}
\usepackage{multirow}
\usepackage[table]{xcolor}
\usepackage{siunitx}
\usepackage{etoolbox}
\usepackage{color}
\usepackage{colortbl,array,xcolor}
\usepackage{booktabs}
\usepackage{pifont}
\newcommand{\cmark}{\ding{51}}
\newcommand{\xmark}{\ding{55}}
\usepackage{tabularx}
\usepackage{makecell}
\usepackage{mathcomp}
\usepackage{array}

\title{ActionCache: Training-Free Acceleration for Vision-Language-Action Models\\with Action Caching and Refinement}

\newcommand{\equal}{$^*$}
\newcommand{\corr}{\textsuperscript{\(\dagger\)}}
\author{%
  \textbf{Ryuji Oi\thanks{Equal contribution.}\hspace{0.5pt}$^,$\corr
  \quad Hikari Otsuka\equal
  \quad Kosuke Matsushima\equal
  \quad Yuki Ichikawa} \\
  \textbf{Masato Motomura
  \quad Tatsuya Kaneko
  \quad Daichi Fujiki}
}

\affiliations{
  Institute of Science Tokyo \\
  \corr oi.ryuji@artic.iir.isct.ac.jp
}

\begin{document}

\maketitle

\begin{abstract}
Vision-Language-Action (VLA) models have emerged as a promising approach for generalizable robotic manipulations. 
In particular, flow-matching-based VLA models have shown remarkable success due to their capability to generate precise and smooth action sequences and capture multimodal distributions. 
However, the iterative denoising process in the action head acts as a major computational bottleneck, posing a critical challenge for real-time deployment. 
To address this challenge, we propose ActionCache, a plug-and-play external cache that opportunistically reuses past intermediate actions to warm-start generations from the vicinity of target actions, drastically reducing the inference latency. 
Specifically, ActionCache stores the intermediate actions with compact multimodal keys, which enables retrieval from similar past contexts across different episodes or even different tasks.
Experimental results in simulation and real-world environments demonstrate that ActionCache maintains high task success rates in a low-latency regime, achieving action head inference acceleration of up to $10.44\times$ and $40.17\times$ for representative flow-based VLA, $\pi_{0.5}$ and GR00T-N1.6, respectively. 
\end{abstract}

\input{1_introduction}
\input{2_background}
\input{3_method}
\input{4_experiment}
\input{5_conclusion}


\bibliography{main}

\newpage

\input{appendix}

\end{document}

%% file: 1_introduction.tex
\section{Introduction}


Vision-Language-Action (VLA) models have emerged as a promising foundation for generalist robot policies that map visual observations and language instructions directly to low-level control~\citep{kawaharazuka2025vla-for-robotics, ma2026vla-survey}.
Many recent VLA models consist of a Vision-Language Model (VLM) backbone and a flow-matching-based action head, which generate continuous action sequences~\citep{shukor2025smolvla, black2026pi0, bjorck2025groot-n1}.
By operating directly in a continuous action space, these flow-based models can avoid action quantization artifacts, capture multimodal action distributions, and produce smooth and precise trajectories.

\begin{figure}[t]
  \centering
  \includegraphics[width=0.99\linewidth]{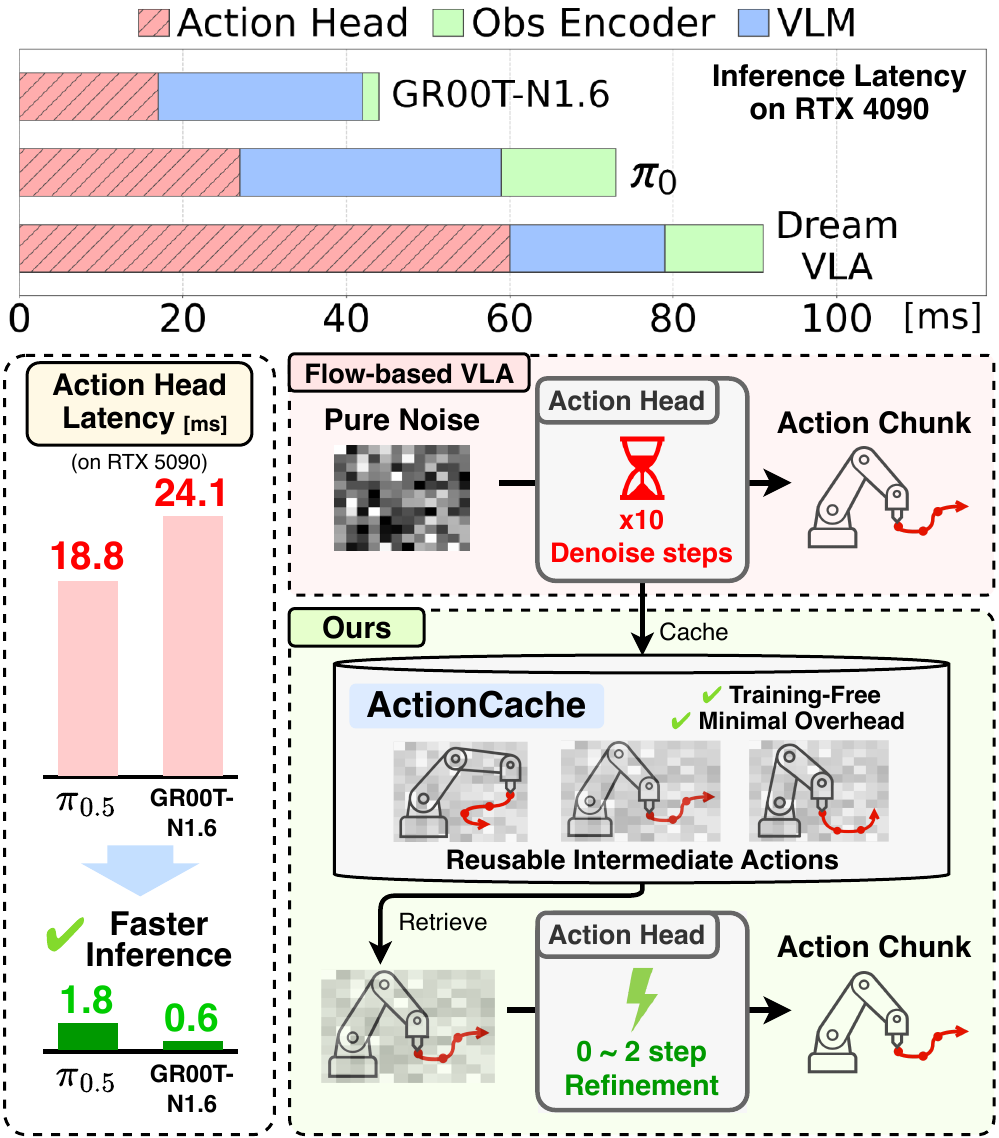}
  \caption{\textbf{Top: Inference latency breakdown on NVIDIA RTX 4090 GPU for representative VLA models.} Action head accountsfor 37-66\% of the end-to-end latency. \textbf{Bottom: ActionCache overview.} ActionCache reduces the inference latency of action head by reusing past actions.}
  \label{figure: action-cache}
\end{figure}

Despite their strong performance, real-time deployment of flow-based VLA models remains challenging because the action head iteratively evaluates a generative model to transform noise into a structured action trajectory at every replanning
Although some prior acceleration methods mainly optimize the VLM backbone~\citep{xu2025vlacache, yang2025efficientvla}, the action head remains a major computational bottleneck, accounting for over 65\% for DreamVLA~\citep{zhang2025dreamvla} and 39\% for GR00T-N1.6~\citep{bjorck2025groot-n1}, as shown in \Cref{figure: action-cache}.

Recently, warm-start methods have been proposed to exploit temporal continuity by initializing refinement from recent actions, predicted actions, or trajectory-level priors~\citep{yufei2025rtidp, li2026step, chen2025falcon, hoeg2025sdp}. 
These methods are typically framed as improving local temporal smoothness by breaking the Markov Decision Process underlying action generation, in which the model is conditioned solely on the current observation. As a byproduct, they can also shorten the effective distance traveled during the iterative refinement by starting closer to the target action trajectory, rather than forcing the model to synthesize an action chunk from noise. 
However, this effect has not been systematically studied as a general mechanism for output reuse beyond temporal continuity. Moreover, existing warm-start methods often depend on learned predictors or explicit temporal modeling rather than offering a plug-and-play solution for pretrained flow-based VLAs.

This paper proposes \textit{ActionCache}, a plug-and-play external cache that turns past action generation into reusable computation. Robotic control often involves recurring visual states, task phases, and language goals, which can induce similar conditional transport paths toward structured action chunks. 
As illustrated in \Cref{figure: action-cache}, ActionCache captures this redundancy by storing generated action chunks with compact multimodal keys and retrieving them when the current context is likely to induce a similar conditional generation path. This reframes warm-starting from a local temporal-continuity heuristic into an output-level retrieval problem. Rather than learning a predictor for the next action or relying solely on previous timesteps, ActionCache opportunistically reuses policy outputs that can initialize refinement closer to the current conditional flow than unstructured noise. In this way, ActionCache substantially reduces the number of required refinement steps in a completely training-free manner, i.e., without retraining the policy, modifying the action head, or adding a learned warm-start module.

Our contributions are as follows:
\begin{itemize}
    \item We introduce ActionCache, a training-free and plug-and-play external cache that accelerates the flow-based action head of VLA models by reusing and refining past action generations.
    \item We develop a model-agnostic cache design that stores intermediate action chunks with compact multimodal keys. It enables reusable computation across timesteps, episodes, and tasks while retaining the pretrained policy through a conservative fallback mechanism.
    \item We validate ActionCache in simulation and real-world robot manipulation tasks, showing substantial improvements in the latency--success-rate trade-off with up to 10.44$\times$ action-head acceleration for $\pi_{0.5}$. We further provide detailed analyses that characterize the factors governing effective action reuse.
\end{itemize}

%% file: 2_background.tex
\section{Background and Related Work}

\subsection{VLA Models}
Vision-Language-Action (VLA) models~\citep{kawaharazuka2025vla-for-robotics, ma2026vla-survey} generate robotic actions based on visual and linguistic inputs.
Most VLA models build on a Vision-Language Model (VLM)~\citep{Zhang2024vlm} backbone and attach an action head that predicts robot actions from the representation extracted by the VLM.
Based on the action generation mechanism of their action heads, VLA models can be categorized into two classes: autoregressive~\citep{brohan2023rt-1, zitkovich2023rt-2, kim2024openvla, li2024roboflamingo, goyal2025vla0} and diffusion-based~\citep{black2026pi0, black2025pi05, bjorck2025groot-n1, shukor2025smolvla, zhang2025dreamvla, wen2025dexvla, li2024cogact, wen2025tinyvla} models.
The former discretizes continuous actions into action tokens and predicts them sequentially, while the latter generates continuous action chunks through an iterative generative process, often using diffusion~\citep{yang2024diffusion-models, ho2020ddpm} or flow matching~\cite{lipman2023flow-matching}.
Many recent VLA models employ diffusion-based action heads, particularly flow-matching-based ones, to model continuous actions more expressively.
These \textit{flow-based} VLAs have achieved remarkable performance across diverse and complex robotic tasks, positioning them as a prominent class of recent VLA models.
Motivated by this trend, this paper focuses on the flow-based VLA architecture family.

At each control timestep $t$, a flow-based VLA passes a context representation $\bm{c}_t$ from the VLM backbone to the flow matching action head.
We note that the exact form of $\bm{c}_t$ is architecture-dependent; it may be obtained from the VLM's final output embeddings~\citep{bjorck2025groot-n1} or from internal representations such as KV caches~\citep{black2026pi0, black2025pi05, shukor2025smolvla}.
Conditioned on $\bm{c}_t$, the action head generates a continuous action chunk by evolving
$\bm{A}_t^\tau := [\bm{a}_t^\tau, \ldots, \bm{a}_{t+H-1}^\tau]$
from $\tau=0$ to $\tau=1$, where $\tau$ is the flow time, and $H$ is the action horizon (the number of actions in the chunk).
Starting from Gaussian noise $\bm{A}_t^0 \sim \mathcal{N}(\bm{0}, \bm{I})$, the action head iteratively transforms the intermediate action chunk as
\begin{equation}
    \bm{A}_t^{\tau + \Delta \tau}
    =
    \bm{A}_t^\tau
    +
    \bm{V}_{\theta}(\bm{A}_t^\tau, \tau, \bm{c}_t) \Delta \tau.
    \label{eq:flow_update}
\end{equation}
Here, $\bm{V}_{\theta}$ is the velocity field predicted by the action head, and $\Delta \tau$ is the numerical integration step size.
After integration to $\tau=1$, the final action chunk is obtained as
$\bm{A}_t^1 = [\bm{a}_t^1, \ldots, \bm{a}_{t+H-1}^1]$.

\subsection{Plug-and-Play Acceleration of Flow-based VLA Models}

\begin{figure*}[t]
  \centering
  \includegraphics[width=0.84\textwidth]{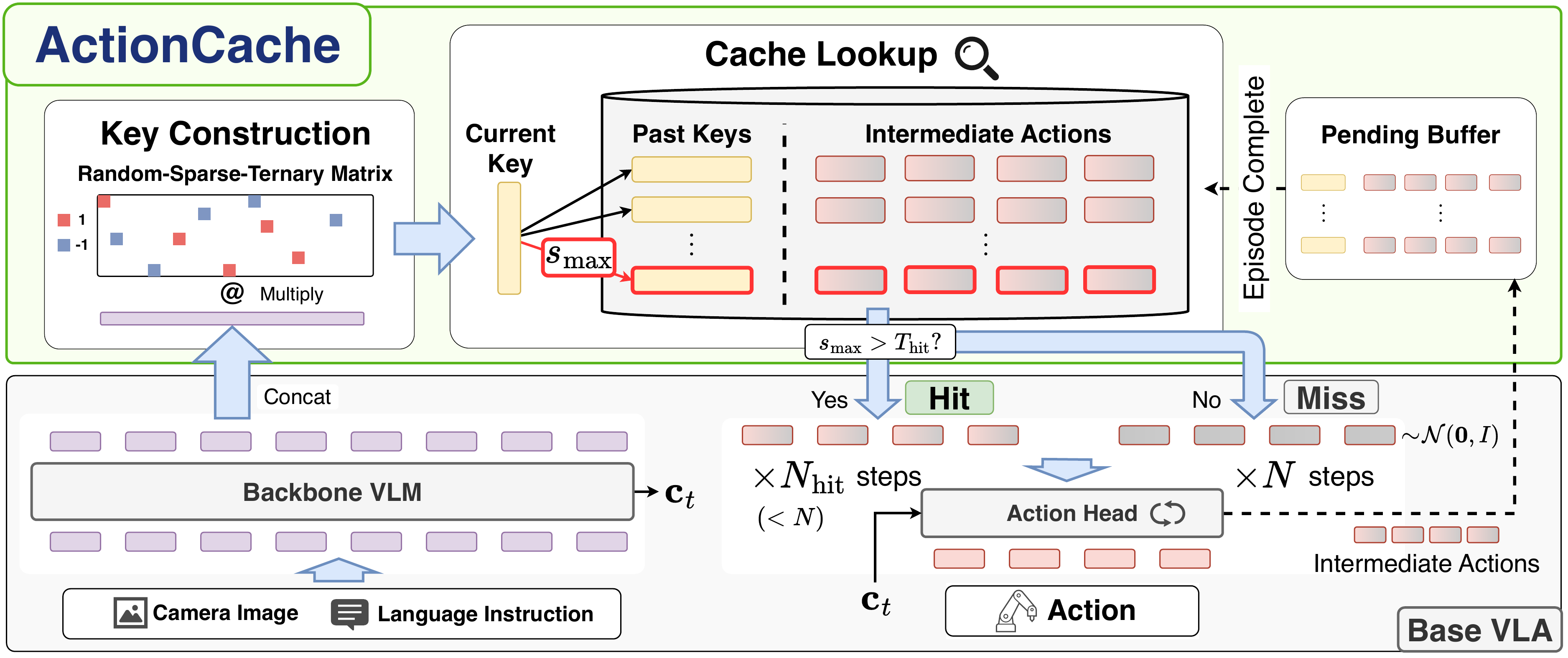}
  \caption{\textbf{ActionCache Framework.} ActionCache is an external cache that stores intermediate noisy actions from past generations. On a cache hit, ActionCache initializes from retrieved past similar actions, reducing the number of denoising steps.}
  \label{figure: action-cache-overview}
\end{figure*}

As shown in \Cref{eq:flow_update}, generating each action chunk in a flow-based VLA requires multi-step flow integration, where the action head repeatedly evaluates the velocity field $\bm{V}_{\theta}$. Thus, the number of flow matching steps directly determines the number of action head forward passes and is a major bottleneck of inference latency.
For instance, in the inference of representative flow-based VLAs, the action head accounts for more than 37\% of the end-to-end latency as shown in \Cref{figure: latency_breakdown}, making it a bottleneck for real-time robot control.

To alleviate this bottleneck, prior works have explored plug-and-play acceleration methods that can be applied to pretrained VLAs without costly retraining.
EfficientVLA~\citep{yang2025efficientvla} reduces redundant computation across the VLA pipeline by pruning less informative language layers, selecting a compact subset of visual tokens, and caching intermediate features in the diffusion-based action head.
While effective, such methods primarily reduce the per-step computational cost or reuse features within the original sampler. 
They do not fundamentally eliminate the need for multi-step flow integration, and the action head must still be invoked repeatedly.
This limitation motivates us to implement a training-free acceleration strategy that more directly reduces the iteration cost of action heads while preserving the behavior of the pretrained VLA.

\subsection{Diffusion Caching on Image Generation}
In image generation, prior work on efficient diffusion inference has shown that caching can reduce denoising costs by reusing previously computed outputs, latent states, or intermediate features.
They retrieve intermediate latent states~\citep{Shubham2024nirvana} or final images~\citep{xia2026modm} of similar previous prompts stored in caches to serve as warm starts, skipping part of the denoising process. Intermediate features across adjacent denoising steps have also been reused to avoid redundant computation~\citep{Mma2024deepcache}.
While these caching methods are training-free and effective for image generation, they cannot be applied directly to flow-based VLA inference, where cache retrieval must be conditioned on multimodal and embodied contexts, including visual observations, language instructions, robot states, and task progress.
To the best of our knowledge, ActionCache is the first method that introduces the caching paradigm to flow-based VLA and optimizes both inference speed and accuracy.

\subsection{Warm-Starting for Diffusion Policies}
Existing warm-starting methods for diffusion-based robot policies~\citep{chi2023diffusionpolicy} mainly exploit local temporal continuity within a rollout by reusing previous action predictions or temporally adjacent trajectories~\citep{li2026step, hoeg2025sdp}.
These methods often need training of predictors or explicit temporal modeling.
STEP~\citep{li2026step} introduces a specialized predictor to generate a temporally and spatially consistent warm-start initialization from the temporally adjacent action trajectory.
SDP~\cite{hoeg2025sdp} trains a diffusion policy to handle action chunks with mixed noise levels, enabling temporal streaming synthesis and accelerated inference.
In contrast, ActionCache aims to reuse action generation across temporally distant but semantically and visually related contexts \emph{without training}.
It enables acceleration by caching action-relevant generation states that are shared across tasks and episodes, rather than relying solely on short-term continuity between consecutive control steps.

%% file: 3_method.tex
\section{Method}

\subsection{Framework Overview}
To address the high inference latency of the iterative action refinement in the flow-based action head, we propose \textit{ActionCache}, a plug-and-play action memoization framework.
\Cref{figure: action-cache-overview} illustrates our framework. 
Given the current observation at a control timestep $t$, the VLM backbone produces a condition $\bm{c}_t$ for the action head.
Leveraging the VLM output embeddings generated as a byproduct of conditioning, ActionCache constructs a cache \emph{key} $\bm{k}_t$, a compact representation that extracts multimodal information from the backbone features in a training-free manner, and uses it to query existing cache entries from past successful action generations. 

Each cache entry contains a cache key $\bm{k}_t$, an action chunk $\bm{A}_t^{\tau}$ generated by the base policy, and optional metadata $\bm{m}_t$ for cache management such as reference counts. 
Importantly, the cached value is the action chunk itself--not internal states, KV cache, or visual token representation--and is therefore defined in the action space rather than tied to a particular backbone architecture. Moreover, it leaves the backbone and action-head weights unchanged. Therefore, ActionCache can be attached to a pretrained VLA policy without additional training.

At inference time, the current key is used to retrieve a candidate cache entry together with a similarity score. ActionCache then branches according to a conservative hit/miss rule. If the similarity of the retrieved entry exceeds the hit threshold $T_\textrm{hit}$, the lookup is treated as reliable, and the retrieved action chunk is used as an initialization for zero- or few-step action generation. Otherwise, the retrieved candidate is rejected, and the model falls back to the standard full-step generation process from pure Gaussian noise $\bm{A}_t^{\tau = 0} \sim \mathcal{N}(\bm{0}, I)$. 
This fallback strategy ensures that the VLA model retains its original robustness and generalization capabilities in unseen or complex situations where the cache is not sufficiently populated.

\subsection{Cache Representation and Retrieval for Multimodal Contexts}\label{subsec:method/cache-rep}
A useful cache key should be expressive enough to capture action-relevant multimodal context while remaining inexpensive to compute and search for at every control step. 
We therefore construct keys from VLM output embeddings produced during the backbone forward pass. 
This choice does not require an additional VLM computation, and in architectures where some VLM output tokens are not directly consumed by the action head, it also reuses features that would otherwise be discarded. 
Moreover, these embeddings encode visual and linguistic context after backbone processing, making them a more action-relevant retrieval signal than raw image or language features alone.
We empirically compare this design with alternative feature sources in Appendix.

Let $\bm{h}_t \in \mathbb{R}^D$ be the concatenated VLM output embeddings. 
Since $D$ can be very large, using $\bm{h}_t$ directly as a cache key would increase memory footprint and lookup cost. 
Therefore, we reduce its dimensionality by projecting it into a compact key 
$\bm{k}_t \in \mathbb{R}^d$ using a fixed sparse ternary random projection matrix~\citep{ping2006sparse-projection}:
\begin{equation}
    \bm{k}_t = \bm{R}\bm{h}_t,
    \qquad
    \bm{R}=(r_{ij}) \in \{-1,0,1\}^{d \times D}.
\end{equation}
Here, each row of $\bm{R}$ contains $pD/2$ entries of 1, $pD/2$ entries of -1, and the remaining $(1-p)D$ entries of 0.
The position of these values is chosen uniformly at random for each row.
$\bm{R}$ is created once and kept fixed.
This compression approach requires no fitting data, retraining, or online update. 
It therefore preserves the plug-and-play nature of ActionCache while reducing storage and similarity search costs. 

Given the current key $\bm{k}_t$, ActionCache retrieves the nearest cache entry $\bm{A}_{i^\star\!}$ using cosine similarity: $i^\star\!=\arg\max_{i} \mathrm{cos\_sim}(\bm{k}_t, \bm{k}_{i})$. 
The retrieved action is considered a hit only when the Top-1 similarity $s_{\max} = \mathrm{cos\_sim}(\bm{k}_t, \bm{k}_{i^\star\!})$ exceeds the threshold $T_{\textrm{hit}}$. On a hit, the retrieved action chunk is either executed directly or used as the initialization for a small number of refinement steps under the current condition $\bm{c}_t$.

\subsection{Cache Population and Management}
During a cache miss, we extract the intermediate noisy action at denoising step $N - N_\textrm{hit}$ from its full generation trajectory.
Here, $N$ denotes the total number of function evaluations (NFE), and $N_\textrm{hit}$ represents the NFE on a cache hit.
This intermediate action $\bm{A}_t^{\tau}$ at $\tau=(N - N_\textrm{hit})\Delta\tau$, paired with its corresponding key $\bm{k}_t$, is temporarily stored in a pending buffer.
If the episode completes successfully, these pending pairs $\{\bm{k}_t, \bm{A}_t^{\tau}\}$ are committed to the main cache for future reuse; otherwise, they are discarded.

Once the cache reaches its maximum capacity, an existing cache entry is evicted and replaced with a new one according to a cache replacement policy. This process allows ActionCache to adapt to changing environments and new tasks. In many routine scenarios, tasks exhibit temporal locality, meaning that related actions are likely to occur again in the near future. This property makes Least Recently Used (LRU) a reasonable policy~\citep{Shubham2024nirvana}. However, LRU can suffer from cache pollution, where a newly inserted but rarely reused entry evicts a more reusable one. This effect becomes more pronounced in a capacity-limited scenario, making Least-Frequently-Used (LFU) a better choice, as we show in the results shortly (\Cref{subsubsec: cache-replacement-policy}). 

ActionCache can be viewed as a generalization of purely temporal warm starting. By modifying the buffering and replacement policy (e.g., to FIFO), it can supply the necessary inputs to a temporal continuity framework, although such work requires an additional learned module to complete warm-starting. ActionCache extends this perspective by introducing an indexed memory that can retrieve action chunks across timesteps, episodes, and tasks. This broader reuse requires cache quality assessment and filtering, but it also allows the policy to benefit from prior successful behavior beyond the immediately preceding control step in a plug-and-play manner.

%% file: 4_experiment.tex
\section{Experiments}\label{sec: experiments}

\robustify\bfseries
\robustify\mdseries
\sisetup{
  text-series-to-math = true
}
\newcommand{\SRunc}[1]{
  \ensuremath{_{\scriptscriptstyle\pm #1}}
}
\newcommand{\SR}[2]{
  #1\SRunc{#2}
}
\newcommand{\bSR}[2]{
  \bfseries #1\mdseries\SRunc{#2}
}
\providecommand{\gray}[1]{#1}
\renewcommand{\gray}[1]{
  \cellcolor{black!6}#1
}
\begin{table*}[t]
  \centering
  \caption{\textbf{Overall performance comparison on VLABench.} ActionCache preserves success rate in low-NFE and low-latency regime, while the simple NFE reduction and prior efficient methods exhibit substantial degradation. The latency of ActionCache includes key projection and cache lookup overhead.}
  \label{tab:comp-sr-latency}
  \small
  \renewcommand{\arraystretch}{0.5}
  \setlength{\tabcolsep}{8pt}
  \begin{tabular}{
    l
    c
    S[
      table-format           = 2.1{\SRunc{3.0}},
      table-align-text-after = false,
      table-model-setup      = \bfseries
    ]
    rr
    c
    S[
      table-format           = 2.1{\SRunc{3.1}},
      table-align-text-after = false,
      table-model-setup      = \bfseries
    ]
    rr
  }
    \toprule[1.1pt]
    &
    \multicolumn{4}{c}{\textbf{$\pi_{0.5}$}}
    &
    \multicolumn{4}{c}{\textbf{GR00T-N1.6}}
    \\
    \cmidrule(lr){2-5}
    \cmidrule(lr){6-9}
    \textbf{Method}
    & \textbf{NFE}
    & {\textbf{SR}($\uparrow$)}
    & \textbf{Lat.}($\downarrow$)
    & \textbf{Speedup}
    & \textbf{NFE}
    & {\textbf{SR}($\uparrow$)}
    & \textbf{Lat.}($\downarrow$)
    & \textbf{Speedup}
    \\
    \midrule[0.7pt]
    \multirow{2}{*}{Base model}
    & $10^\dagger$
    & \SR{38.8}{3.0}
    & 18.8
    & 1.00$\times$
    & $4^\dagger$
    & \bSR{34.0}{1.2}
    & 24.1
    & 1.00$\times$
    \\

    & 1
    & \SR{6.8}{2.9}
    & 2.5
    & 7.52$\times$
    & 1
    & \SR{24.6}{1.4}
    & 6.9
    & 3.49$\times$
    \\
    \midrule[0.5pt]
    \multirow{2}{*}{EfficientVLA}
    & 2
    & 17.9
    & 5.2
    & 3.62$\times$
    & 2
    & 19.7
    & 15.3
    & 1.58$\times$
    \\
    & 1
    & 7.3
    & 3.6
    & 5.22$\times$
    & 1
    & 8.2
    & 9.2
    & 2.62$\times$
    \\
    \midrule[0.5pt]
    \multirow{2}{*}{Falcon}
    & --
    & 13.1
    & 13.0
    & 1.45$\times$
    & --
    & 26.1
    & 18.8
    & 1.28$\times$
    \\
    & --
    & 7.6
    & 6.8
    & 2.76$\times$
    & --
    & 19.3
    & 15.6
    & 1.54$\times$
    \\
    \specialrule{\lightrulewidth}{\aboverulesep}{0.15ex}
    \rowcolor{black!6}
    \rule{0pt}{2ex}
    & 2
    & \SR{39.4}{2.8}
    & 5.5
    & 3.42$\times$
    & 2
    & \SR{33.9}{0.9}
    & 13.6
    & 1.77$\times$
    \\
    \rowcolor{black!6}
    & 1
    & \bSR{41.0}{1.2}
    & 3.6
    & 5.22$\times$
    & 1
    & \SR{32.1}{3.2}
    & 7.5
    & 3.21$\times$
    \\
    \rowcolor{black!6}
    \multirow{-4.5}{*}{\textbf{ActionCache (Ours)}}
    & 0
    & \SR{40.9}{2.8}
    & \textbf{1.8}
    & \textbf{10.44$\times$}
    & 0
    & \SR{30.8}{1.8}
    & \textbf{0.6}
    & \textbf{40.17$\times$}
    \\
    \specialrule{1.1pt}{0.5pt}{0pt}
    \multicolumn{9}{@{}l}{%
      \small
      $^\dagger$ Default NFE for each base model. SR(\%) and Lat.(ms) denote success rate and inference latency, respectively.
    }
  \end{tabular}
\end{table*}

\begin{figure*}[t]
    \centering
    \includegraphics[width=0.55\textwidth]{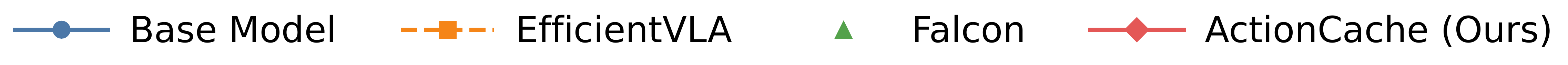} 
    \\ 
    \includegraphics[width=0.99\textwidth]{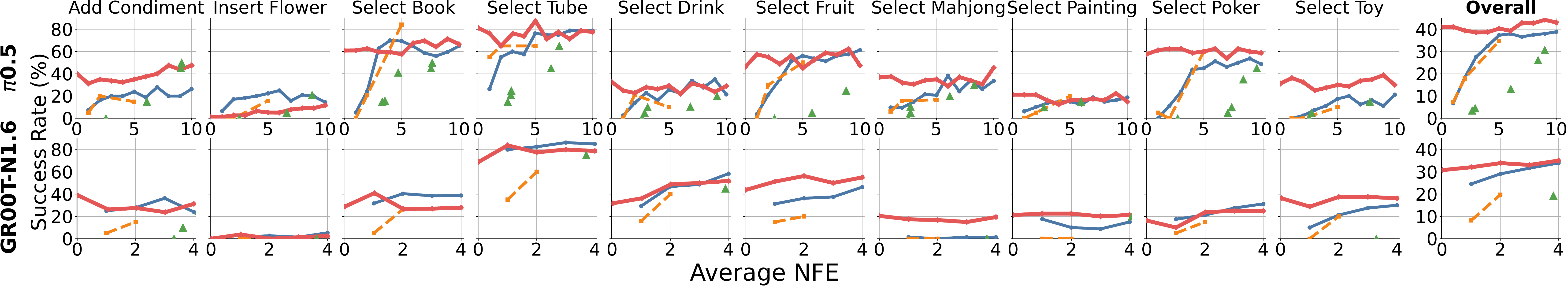}
    \caption{\textbf{Success rate and average NFEs on VLABench.} ActionCache provides a better trade-off of NFE and success rate. 
    }
    \label{figure: avgnfe-vs-sr-per-task}
\end{figure*}

\begin{figure*}[t]
  \centering
  \begin{subfigure}[b]{0.33\textwidth}
    \centering
    \includegraphics[width=\textwidth]{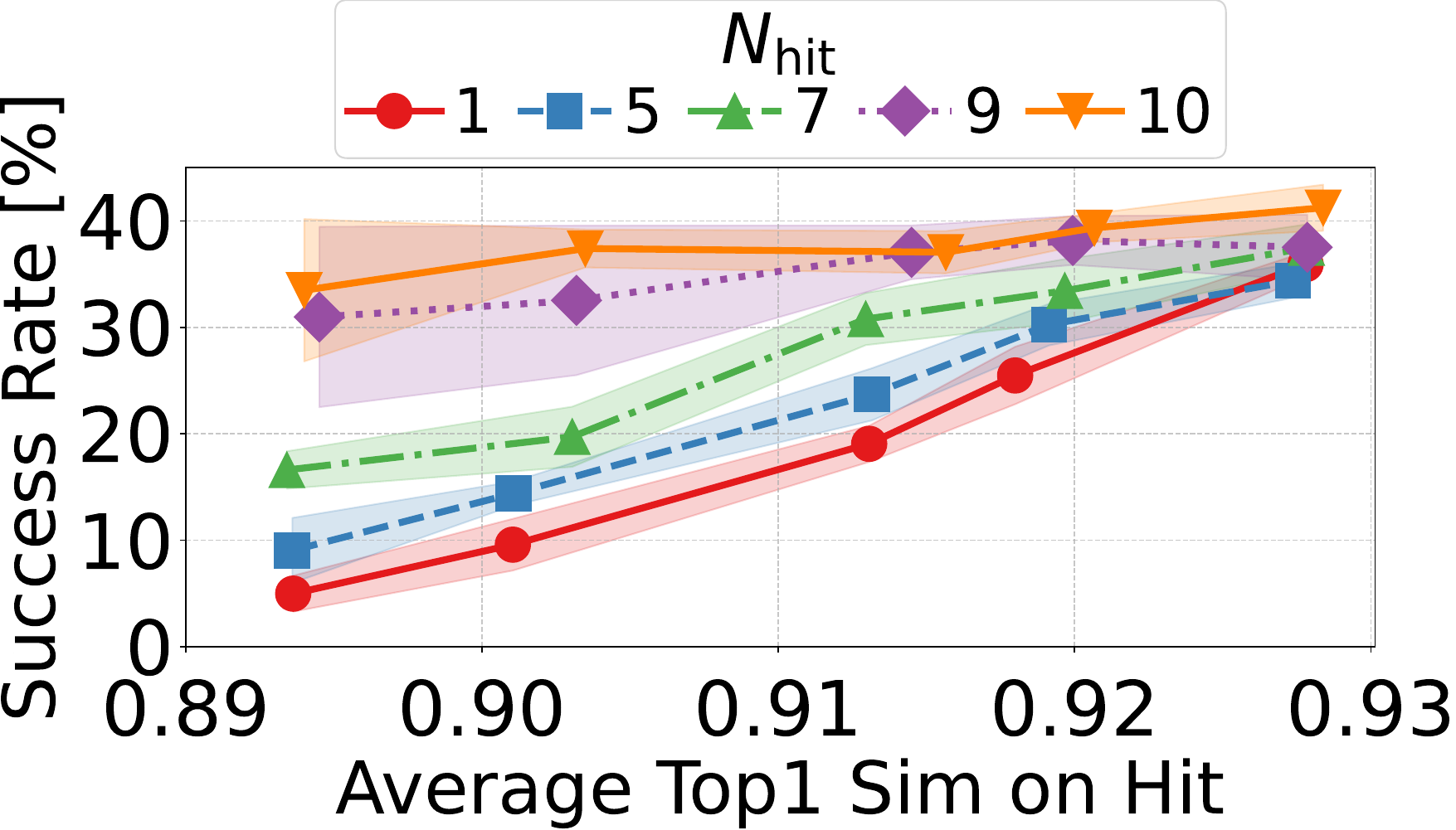}
    \caption{}
    \label{figure: avgtop1-vs-sr}
  \end{subfigure}
  \hfill 
  \begin{subfigure}[b]{0.33\textwidth}
    \centering
    \includegraphics[width=\textwidth]{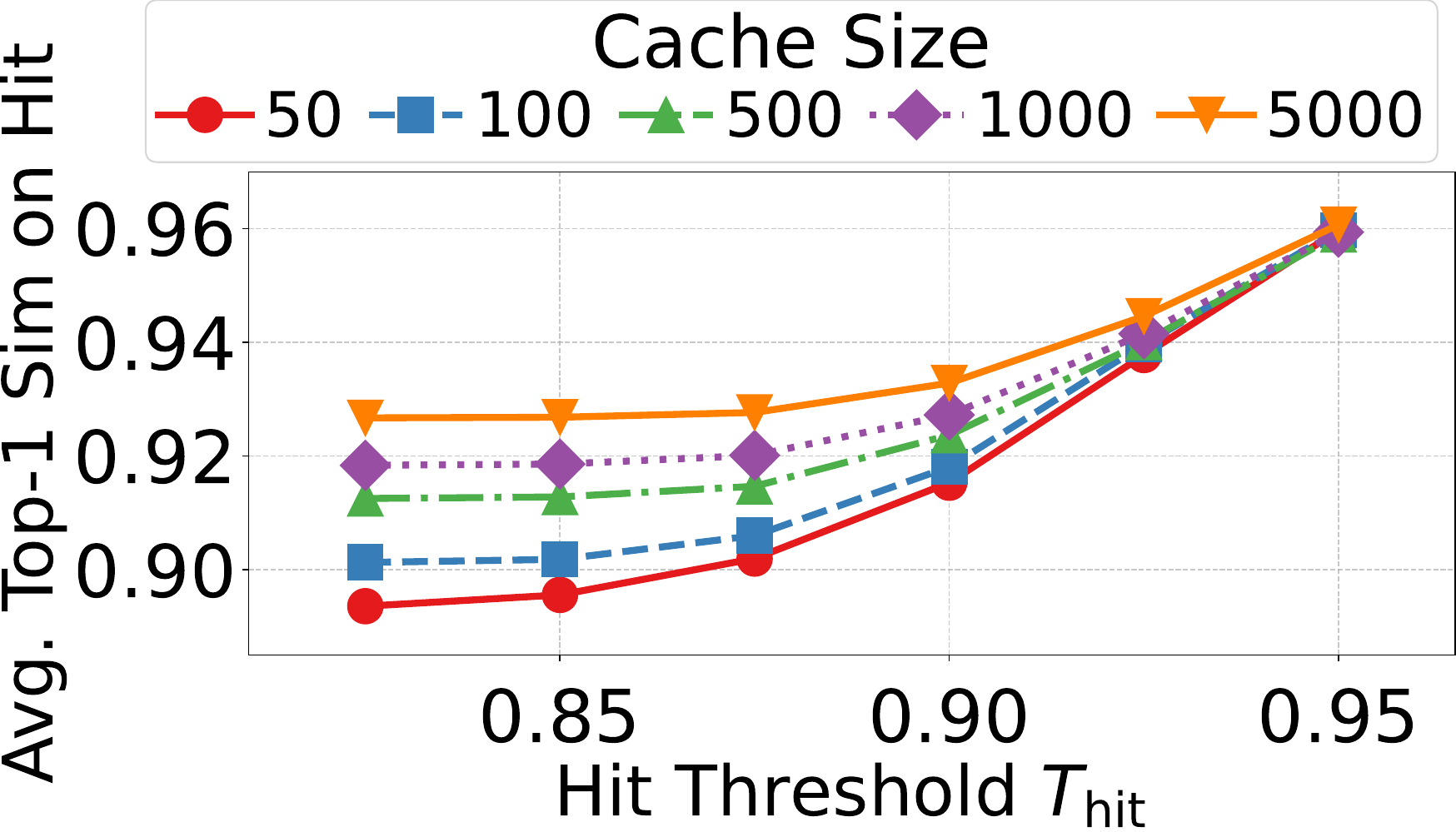}
    \caption{}
    \label{figure: thres-vs-top1sim}
  \end{subfigure}
  \hfill 
  \begin{subfigure}[b]{0.33\textwidth}
    \centering
    \includegraphics[width=\textwidth]{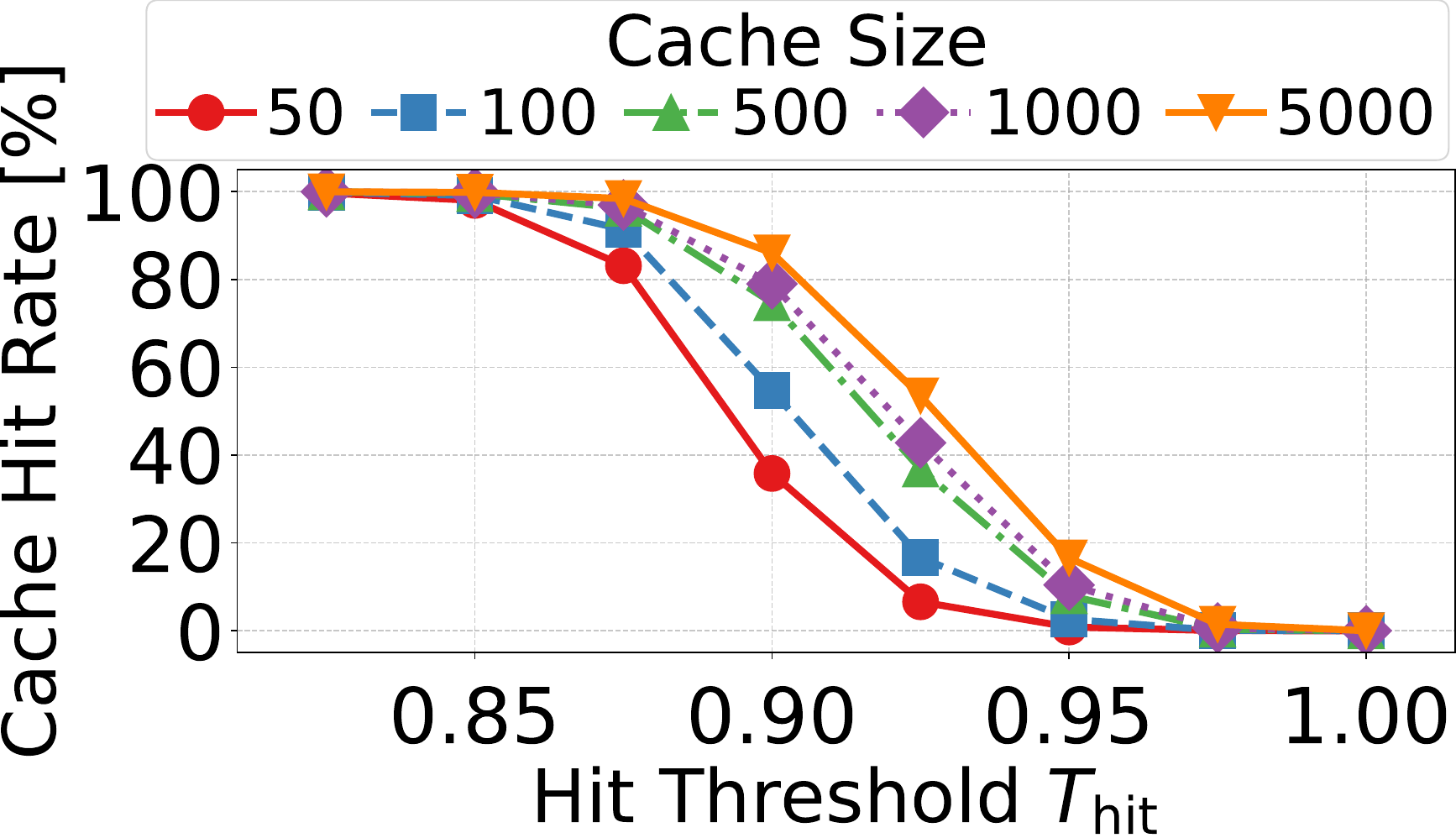}
    \caption{}
    \label{figure: th-vs-chr-cs}
  \end{subfigure}
  \caption{\textbf{(a): Average Top-1 similarity vs success rate for various numbers of denoising steps.} As the Top-1 similarity increases, the success rates improve. \textbf{(b): Average Top1 similarity vs hit thresholds for various cache sizes.} As the hit threshold increases, average Top-1 similarity also increases. \textbf{(c): Cache hit rate vs various hit thresholds for various cache sizes.} A Larger cache yields a higher cache hit rate.}
\end{figure*}

\subsection{Evaluation Settings}\label{subsec: eval-setting}

\textbf{Implementation details.}
We evaluate two state-of-the-art flow-based VLA models: $\pi_{0.5}$~\citep{black2025pi05} and GR00T-N1.6~\citep{bjorck2025groot-n1}.
For GR00T-N1.6, instead of raw VLM output embeddings, which do not include robot-state features, we use a concatenation of encoded VLM output embeddings and encoded robot-state features as a source of cache key.
We set the dimension of the cache key to $d = 500$ with the non-zero value density of sparse ternary random matrix $p = 0.01$.
Since $d$ and $p$ are insensitive to success rate (see Appendix), we select sufficiently large values that can maintain efficient cache lookup.
To evaluate steady-state performance of ActionCache, we fill the cache by running the rollouts with $T_\textrm{hit}=1$, and then start evaluation with a fixed $T_\textrm{hit}$.
$T_\textrm{hit}$ is systematically selected by profiling the similarity distribution of the prefill phase and identifying the shoulder of the hit-rate curve before it plateaus. By default, $T_\textrm{hit}$ is set to $0.85$ for $\pi_{0.5}$ and $0.65$ for GR00T-N1.6.
Prefill and test episodes use disjoint random seeds, ensuring no cached action is generated under an identical task configuration (e.g., object placement) to any test episode.
The default cache size is set to $3{,}000$ entries managed by the LRU policy.

For both models, we set the action execution horizon to $10$.
Unless otherwise specified, we use these hyperparameters as the default configuration throughout our experiments.

\textbf{Simulation Evaluation.}
To validate the efficacy of our framework in various task settings, we use VLABench~\citep{zhang2024vlabench} and LIBERO~\citep{liu2023libero} benchmarks.
VLABench is a large-scale robot manipulation benchmark that contains $100$ tasks with $2,000+$ objects and evaluates visual/spatial understanding and common-sense/world-knowledge application.
We evaluate on the 10 primitive tasks of varying difficulty from VLABench, which are released as an official unified dataset on Hugging Face.
Throughout prefill and test time, these tasks are executed in a round-robin manner, which means the cache contains entries from all the tasks.
The latency is measured on a single NVIDIA RTX 5090 GPU with $32$ GB of GPU memory. 
More details and the results for the LIBERO benchmark are reported in Appendix.

\textbf{Real-World Evaluation.} 
We deploy the $\pi_{0.5}$ model on SO-101~\citep{cadene2026lerobot}, an open-sourced 6-DoF robotic arm.
We evaluate on three tasks: \texttt{pick\_sausage}, \texttt{pick\_and\_close}, and \texttt{push\_button}.
These tasks cover evaluation of robustness to distractor objects and spatial perturbations, multi-step long-horizon manipulation, and instruction following through visual-language grounding. Detailed task descriptions and fine-tuning settings are provided in the Appendix.

\subsection{Success Rate and Latency Evaluation}\label{subsec: eval-sr-latency}

\textbf{Overall Performance Comparison on VLABench.}
To validate the efficacy of ActionCache, we compare with a simple denoising step reduction model and two prior training-free acceleration methods, EfficientVLA~\citep{yang2025efficientvla} and Falcon~\citep{chen2025falcon}.
We measure success rate with a total of $800$ episodes for base models and ActionCache, and $200$ episodes for other methods.
We set the cache size to $10{,}000$ for $\pi_{0.5}$ and $30{,}000$ for GR00T-N1.6, and evaluate its steady-state performance.
\Cref{tab:comp-sr-latency} summarizes the number of function evaluations (NFE), success rate and inference latency of the action head in the low-latency (low-NFE) regime.
Note that the NFE of ActionCache is equal to the $N_\textrm{hit}$ value.

Although the base models drop their success rates severely at $\mathrm{NFE}=1$, ActionCache mitigates the degradation in the low-NFE regime.
This result demonstrates that by starting from the vicinity of the target action rather than from pure noise, ActionCache can generate a high-fidelity action even with a minimal NFE.
Compared to EfficientVLA, which is a prior plug-and-play acceleration method, ActionCache achieves a better trade-off between success rate and latency.
This is because EfficientVLA's acceleration is limited by its reliance on pure noise initialization and its inability to skip the entire execution of a denoising step.
Falcon, a warm-starting method for diffusion policies~\citep{chi2023diffusionpolicy}, exhibits a less favorable latency--success-rate trade-off likely due to the overhead of selecting an initial trajectory and a limited correctionability of recent-trajectory warm-starting under deterministic flow integration.
Furthermore, especially on $\pi_{0.5}$, ActionCache maintains a high success rate at $\mathrm{NFE}=0$ with negligible latency overhead induced by key projection and cache lookup.
By directly reusing past actions in similar contexts, ActionCache achieves a drastic speedup, suggesting that past actions can be retrieved and reused without modification.

\Cref{figure: avgnfe-vs-sr-per-task} shows the trade-off between average NFE and success rate for each task in VLABench.
Overall, ActionCache achieves a better trade-off compared to the base model and the other prior methods across a broad range of tasks, especially in the low-NFE regime.

%

\begin{figure}[h]
  \centering
  \small
  \makeatletter
  \def\@captype{table} 
  \makeatother
  \setlength{\tabcolsep}{8pt}
  \renewcommand{\arraystretch}{0.7}
  \begin{tabular}{lc}
    \toprule
    Target Task & Progress Score (\%) \\
    \midrule
    \texttt{select\_painting} & 20.0 (base: 21.0) \\
    \texttt{select\_toy}      & 51.7 (base: 50.5) \\
    \bottomrule
  \end{tabular}

  \makeatletter
  \def\@captype{figure} 
  \makeatother
  \includegraphics[width=0.35\textwidth]{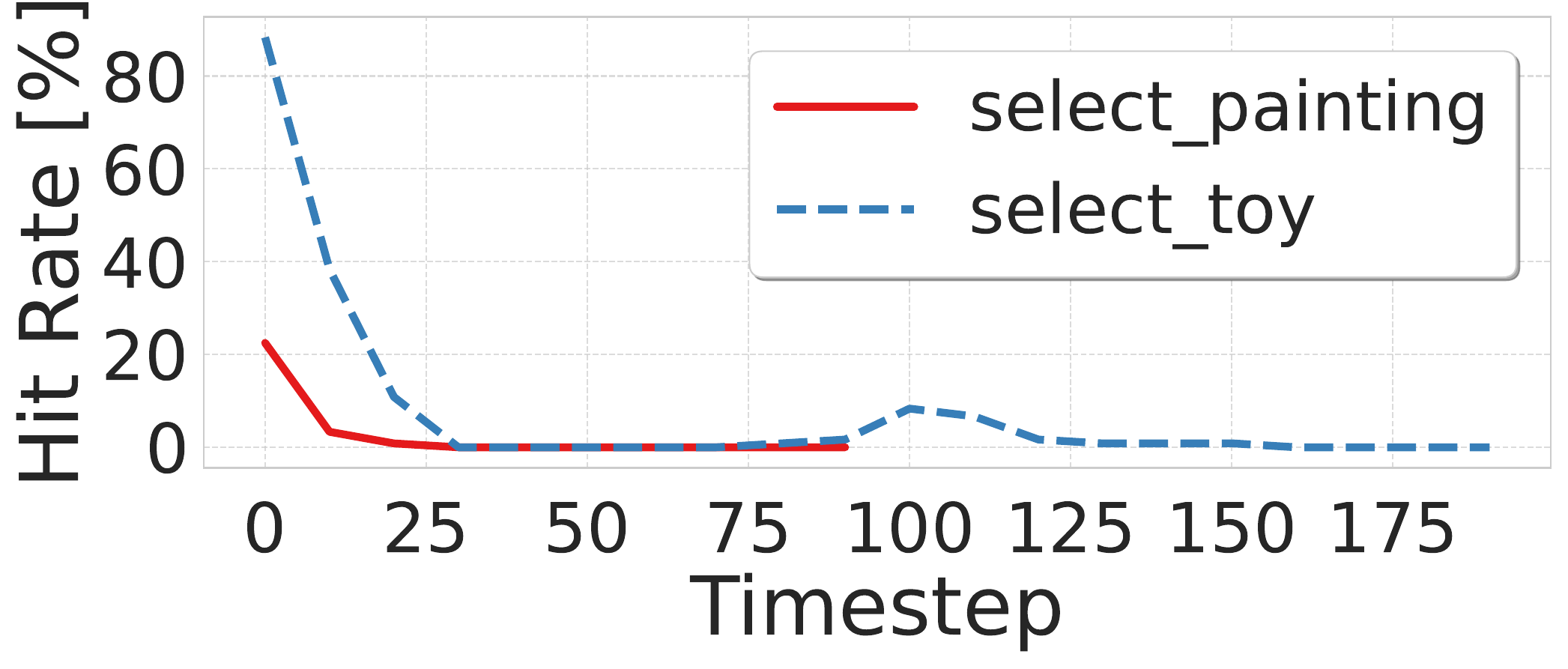}
  \caption{\textbf{Progress score on the target tasks and cache hit rate across rollout timesteps.} ActionCache maintains the base model's progress scores in unseen tasks. Cache hit rate reaches over 80\% at the starting stage.}
  \label{figure:cross-task-step-vs-chr}
\end{figure}

\subsection{Cross-Task Action Reuse} \label{subsec: cross-task-reuse}
To investigate the potential for cross-task action reuse, we conduct a zero-shot cross-task experiment on VLABench with $\pi_{0.5}$.
In this experiment, we first populate the cache with the \texttt{select\_fruit} task and then evaluate the model's performance on other target tasks.
For more fine-grained analysis of action reuse across different tasks, we use a task progress score for evaluation, which reflects the completion of intermediate sub-goals within a task, rather than a binary success rate.
During the evaluation of the target task, we disabled cache updates to ensure that any hits represent cross-task action sharing.
The hit threshold $T_\textrm{hit}$ is set to 0.925 during the target task evaluation. 
Results are aggregated over 120 episodes for each task.

\textbf{Progress Score and Hit Rate.}
\Cref{figure:cross-task-step-vs-chr} shows that ActionCache maintains the base model's progress score while providing a meaningful cache hit rate at a certain phase. 
These results demonstrate that the model can retrieve and reuse actions from different tasks to make meaningful progress on unseen tasks, while safely falling back to the original model when uncertain or unpopulated, demonstrating both cross-task effectiveness and robustness.
Notably, the hit rate reaches over $80$\% at the initial stage of episodes.
This result suggests that the early stages of episodes, such as reaching motions toward a target object, share fundamental actions across multiple tasks, revealing the potential for action reuse beyond individual task boundaries.

\begin{table*}[t]
  \centering
  \caption{\textbf{Success rate and latency breakdown of $\pi_{0.5}$ in the real-world environment.} ActionCache maintains success rate while reducing action head latency with a negligible overhead. SR and HR denote success rate and hit rate, respectively.}
  \label{tab:latency-vs-sr-real}
  \small
  \renewcommand{\arraystretch}{0.5}
  \setlength{\tabcolsep}{4pt}
  \begin{tabular}{llcccccccr | r}
    \toprule
    \multirow[b]{2}{*}{Task}
      & \multirow[b]{2}{*}{Method}
      & \multirow[b]{2}{*}{NFE}
      & \multirow[b]{2}{*}{\begin{tabular}[c]{@{}c@{}}SR\\(\%)\end{tabular}}
      & \multirow[b]{2}{*}{\begin{tabular}[c]{@{}c@{}}HR\\(\%)\end{tabular}}
      & \multicolumn{6}{c}{Latency (ms)} \\
    \cmidrule(lr){6-11}
      & & & & & Emb. & VLM & Key proj. & Cache lookup & Action head
      & \multicolumn{1}{c}{Overall} \\
    \midrule

    \multirow[c]{2}{*}{\texttt{pick\_sausage}}
      & Base Model
      & 10 & 90 & -
      & 24.22 & 22.21 & - & -
      & 56.70 (1.00$\times$)
      & 103.13 (1.00$\times$) \\

      & \textbf{ActionCache}
      & 1 & 88 & 41.8
      & 24.26 & 22.24 & 0.17 & 0.18
      & 35.06 (\textbf{1.62$\times$})
      & 81.90 (\textbf{1.26$\times$}) \\

    \midrule

    \multirow[c]{2}{*}{\texttt{pick\_and\_close}}
      & Base Model
      & 10 & 88 & -
      & 24.23 & 22.22 & - & -
      & 55.79 (1.00$\times$)
      & 102.24 (1.00$\times$) \\

      & \textbf{ActionCache}
      & 1 & 90 & 82.8
      & 24.29 & 22.26 & 0.17 & 0.17
      & 14.67 (\textbf{3.80$\times$})
      & 61.56 (\textbf{1.66$\times$}) \\

      \midrule

    \multirow[c]{2}{*}{\texttt{push\_button}}
      & Base Model
      & 10 & 100 & -
      & 24.31 & 22.29 & - & -
      & 59.65 (1.00$\times$)
      & 106.26 (1.00$\times$) \\

      & \textbf{ActionCache}
      & 1 & 100 & 94.4
      & 24.33 & 22.32 & 0.17 & 0.11
      & 9.53 (\textbf{6.26$\times$})
      & 56.46 (\textbf{1.88$\times$}) \\

    \bottomrule
  \end{tabular}
\end{table*}

\subsection{Impact of Retrieval Quality on Success Rate}

\textbf{Retrieval Quality and Success Rate.}\label{subsec: ret-quality-vs-sccess-rate}
We investigate the relationship between the quality of retrieved actions and task success rate with varying hyperparameters for ActionCache.
The quality of retrieved actions is measured by average Top-1 cosine similarity on cache hits, which we control by varying the cache size from 50 (114 KB) to 5,000 (11.4 MB).

From \Cref{figure: avgtop1-vs-sr}, we can observe that when the average Top-1 similarity on cache hits is high enough, the model achieves high success rates comparable to that of the full-step generation, regardless of $N_\textrm{hit}$. 
This observation indicates that ActionCache requires either a sufficiently large $N_\textrm{hit}$ or a high average Top-1 similarity to achieve a high task success rate.
Since increasing $N_\textrm{hit}$ negates the computational benefits of ActionCache, it is necessary to increase the Top-1 similarity on retrieval for a good trade-off between success rate and latency.

Hit threshold $T_\textrm{hit}$ serves as a key parameter to control success rate.
\Cref{figure: thres-vs-top1sim} shows the relationship between the average Top-1 similarity and $T_\textrm{hit}$, where a higher $T_\textrm{hit}$ leads to a higher average Top-1 similarity.
With a sufficiently large cache, the average Top-1 similarity saturates at lower thresholds because the nearest-neighbor similarity distribution is already concentrated above them, making the hit criterion non-selective.
Since $T_\textrm{hit}$ enforces a lower bound on the similarity of accepted retrievals, it provides a practical control knob for the quality of retrieved actions and, consequently, the expected success rate.
This allows us to tune the trade-off of cache capacity and latency benefit as below with a target success rate.

Finally, \Cref{figure: th-vs-chr-cs} shows the cache hit rate with varying $T_\textrm{hit}$ and cache sizes. 
We observe that increasing the cache size monotonically improves the cache hit rate, which is directly translated into lower latency.
Overall, these results indicate that ActionCache provides highly flexible knobs to adjust accuracy-efficiency-capacity trade-offs, depending on the task difficulty, success rate, target latency, and allocated memory.

\begin{figure}[t]
  \centering
  \includegraphics[width=0.98\columnwidth]{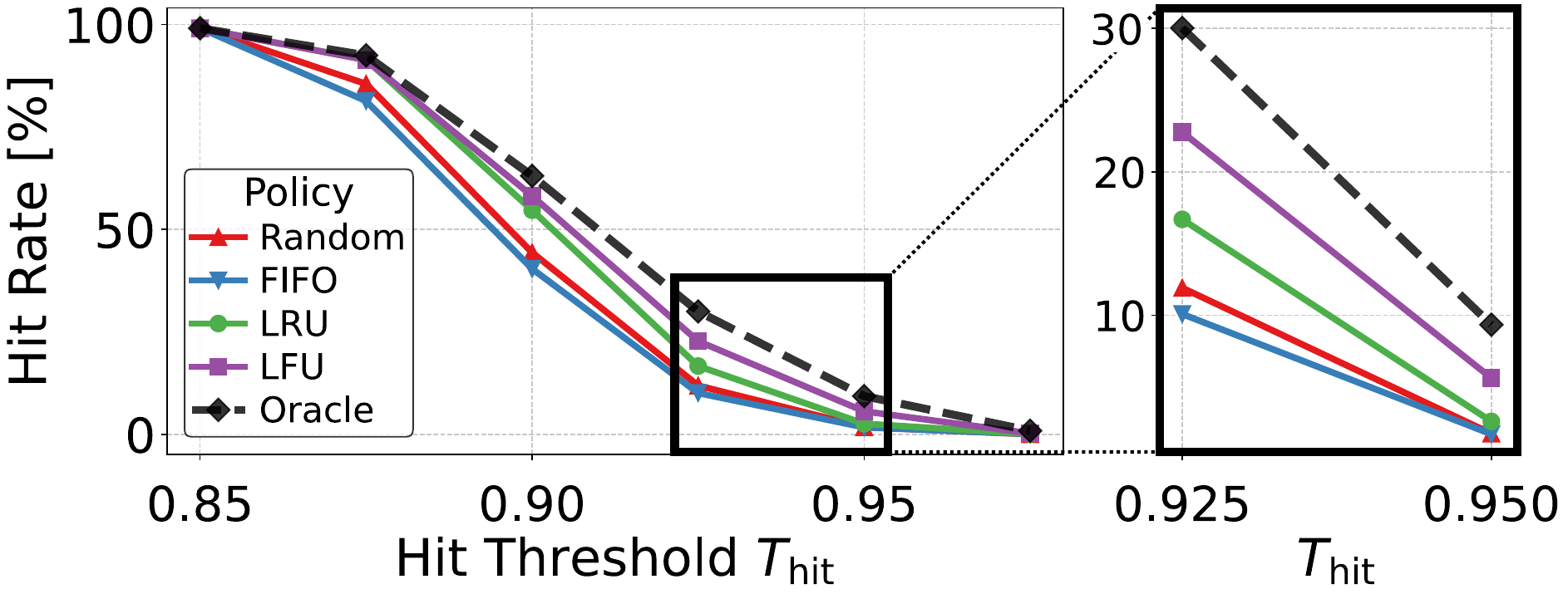}
  \caption{\textbf{Hit rate comparison for cache replacement policies under a restricted cache size of 100.} LFU policy demonstrates the highest hit rate across hit thresholds.}
  \label{figure: th-vs-chr-policy}
\end{figure}

\textbf{Comparison of Cache Replacement Policies.}\label{subsubsec: cache-replacement-policy}
As established in \Cref{subsec: ret-quality-vs-sccess-rate}, the quality of retrieved actions is important to achieve a high success rate in ActionCache.
With a large cache size, we can retrieve sufficiently high-quality actions from the cache with a higher cache hit rate, whereas in a restricted cache size, the replacement policy is crucial for a higher cache hit rate under a specific hit threshold $T_\textrm{hit}$.
To this end, we investigate the impact of different cache replacement policies on the cache hit rate.
We compare five policies: Random, Least Recently Used (LRU), Least Frequently Used (LFU), First In First Out (FIFO), and an oracle algorithm~\citep{belady}, which assumes future knowledge of queries and serves as the ideal replacement policy.

As shown in \Cref{figure: th-vs-chr-policy}, in limited capacity, the LFU policy demonstrates the highest cache hit rate among the other policies.
In a typical caching mechanism, the LRU policy is likely to achieve a high hit rate due to its ability to exploit temporal locality.
While LRU is susceptible to cache pollution by \emph{one-hit wonder} actions, such as highly specific and difficult grasps, LFU prioritizes robust and highly reusable actions that are shared across tasks or episodes. Although LFU captures temporal locality less effectively than LRU, retaining task- or episode-agnostic highly reusable actions in the cache contributes greatly to the hit rate in those settings.
Notably, the FIFO policy, which implicitly measures the probability of encountering a good starting point (specified by $T_\textrm{hit}$) of traditional warm-starting methods for temporal continuity, is no better than random replacement in a plug-and-play setting, suggesting a missed opportunity for caching.





\subsection{Combination with prior VLM acceleration}\label{subsec: vla-cache}

\begin{table}[t]
    \centering
    \small
    \caption{\textbf{End-to-end $\pi_{0.5}$ performance comparison.}}
    \label{tab: vla-cache}
    \setlength{\tabcolsep}{5pt}
    \renewcommand{\arraystretch}{0.8}

    \begin{tabular}{cc|cccc}
        \toprule

        \multirow{2}{*}{%
            \textbf{\shortstack[c]{VLA-\\Cache}}}
        &
        \multirow{2}{*}{%
            \textbf{\shortstack[c]{Action\\Cache}}}
        &
        \multirow{2}{*}{%
            \textbf{\shortstack[c]{SR\\(\%)}}}
        &
        \multicolumn{3}{c}{\textbf{Latency (ms)}}
        \\

        \cmidrule(lr){4-6}

        &
        &
        &
        \textbf{VLM}
        &
        \textbf{Act. Head}
        &
        \textbf{Overall}
        \\

        \midrule

        \xmark
        & \xmark
        & 40.7
        & 27.6
        & 58.4
        & 97.4 (1.00$\times$)
        \\


        \cmark
        & \xmark
        & \textbf{41.4}
        & 24.0
        & 58.8
        & 94.2 (1.03$\times$)
        \\

        \xmark
        & \cmark
        & 41.2
        & 27.6
        & \textbf{6.8}
        & \textbf{45.9} (\textbf{2.12$\times$)}
        \\


        \cmark
        & \cmark
        & 39.1
        & \textbf{23.9}
        & 21.6
        & 57.0 (1.71$\times$)
        \\

        \bottomrule
    \end{tabular}
\end{table}

To measure end-to-end performance, we evaluate success rate and latency by coupling ActionCache with VLA-Cache, a prior \emph{VLM backbone acceleration method} which reduces redundant computation by adaptively reusing visual token KV cache.
For this experiment, we use $\pi_{0.5}$ and select three tasks with varying difficulty from VLABench: \texttt{select\_tube}, \texttt{select\_painting} and \texttt{add\_condiment}.
For ActionCache, the cache size is set to $3{,}000$.
More details, including VLA-Cache settings and implementation, are provided in Appendix.
As shown in \Cref{tab: vla-cache}, ActionCache works seamlessly with VLA-Cache, maintaining its success rate, while ActionCache requires a stricter $T_\textrm{hit}$ due to the similarity distribution distorted by VLA-Cache's cached image tokens.
This result indicates that our plug-and-play framework can be combined with VLM acceleration methods without compromising performance, realizing end-to-end acceleration.

\subsection{Evaluation on Real-World}\label{subsec: eval-real-world}

To validate the efficacy of ActionCache in a real-world environment, we evaluate $\pi_{0.5}$ on three robotic arm manipulation tasks.
For ActionCache, we set the cache size to $300$ for the \texttt{push\_button} task and $1{,}000$ for the others.
$T_\textrm{hit}$ is set to $0.875$ for \texttt{pick\_sausage} and $0.85$ for the others. 

\textbf{Success Rate and Inference Latency.}
\Cref{tab:latency-vs-sr-real} summarizes the results.
Overall, ActionCache achieves a success rate comparable to the base model while reducing action-head latency, improving overall latency, and demonstrating practical applicability in real-world settings.
\Cref{tab:latency-vs-sr-real} also shows that ActionCache can be introduced with only a negligible latency overhead.
Notably, the \texttt{push\_button} task achieves over 94\% hit rate while maintaining a 100\% success rate.
Despite its longer horizon resulting in a smaller number of cached episodes, the \texttt{pick\_and\_close} task exhibits a higher cache hit rate than the \texttt{pick\_sausage} task.
Since the \texttt{pick\_sausage} includes randomly positioned task-irrelevant distractor objects, this result suggests that visual clutter and observation perturbations might affect the key matching process.
More detailed analyses about key similarity are provided in Appendix.



%% file: 5_conclusion.tex
\section{Conclusion and Limitations}

In this paper, we propose ActionCache, a plug-and-play external cache that accelerates action heads, a major bottleneck in flow-based VLA inference.
Through our experiments, we demonstrate that ActionCache can drastically improve the latency--success-rate trade-off of action head compared to the base model and prior acceleration methods.

\textbf{Limitations.} 
Our framework introduces extra hyperparameters, $T_\textrm{hit}$ and $N_\textrm{hit}$, in addition to accuracy insensitive ones  $d$ and $p$ (see Appendix). While $T_\textrm{hit}$ and $N_\textrm{hit}$ can be tuned before deployment based on training/prefill information and our hyperparameter analysis in \Cref{subsec: ret-quality-vs-sccess-rate}, further optimization and dynamic adaptation of these may improve the trade-off between accuracy and acceleration. We leave the design of effective hyperparameter optimization strategies for future work.

%% file: appendix.tex
\renewcommand{\thesection}{\Alph{section}}
\renewcommand{\thesubsection}{\thesection.\arabic{subsection}}
\setcounter{secnumdepth}{2} 
\setcounter{section}{0}

\onecolumn
\section*{\huge Appendix}
\input{app_key_source}
\input{app_impl_detail}
\input{app_hyperparam}

\input{app_longspan}
\input{app_similarity}
\input{app_libero}

%% file: app_key_source.tex
\section{Ablation Study on Cache Key Source}\label{sec: ablation}

\begin{table}[h]
  \centering
  \small
  \caption{\textbf{Ablation of Cache Key Source.} Using VLM output embeddings as a cache key source outperforms the VLM input embeddings across various cache hit rates.}
  \label{tab: key-source-comp}
  \renewcommand{\arraystretch}{0.8}
  \newcommand{\meanstd}[2]{#1_{\scriptscriptstyle \pm #2}}
  \setlength{\tabcolsep}{5pt}
  \begin{tabular}{l ccc ccc}
    \toprule[1.2pt]
    Source & \multicolumn{3}{c}{VLM-in} & \multicolumn{3}{c}{\textbf{VLM-out}} \\
    \cmidrule(lr){2-4} \cmidrule(lr){5-7}
    HR($\%$)
    & 99.9
    & 83.0
    & 30.0
    & 99.8
    & 83.7
    & 17.9
    \\
    \midrule[0.5pt]

    SR($\%$)
    & $\meanstd{7.3}{0.8}$
    & $\meanstd{8.1}{1.9}$
    & $\meanstd{19.7}{1.7}$
    & $\meanstd{\textbf{34.7}}{3.9}$
    & $\meanstd{\textbf{37.2}}{3.2}$
    & $\meanstd{\textbf{39.7}}{3.1}$
    \\

    \bottomrule[1.2pt]
  \end{tabular}
\end{table}

ActionCache utilizes output embeddings of the VLM as an action-relevant cache key source.
To validate this design choice, we investigate how the source of cache key impacts overall success rate on VLABench.
As shown in \Cref{tab: key-source-comp}, utilizing VLM output embeddings significantly outperforms the VLM input counterpart by up to $29.1$\% under roughly the same cache hit rate.
This performance gap indicates that VLM outputs more effectively encode action-relevant multimodal context than VLM inputs.

%% file: app_impl_detail.tex
\section{Implementation Details for Simulation/Real-world Evaluations}\label[appendix]{appsec:implementation-details}

\subsection{Model Checkpoints}
We utilized some publicly available model checkpoints for our experiments.
For $\pi_{0.5}$, we used \texttt{VLABench/pi05-primitive-10task}\footnote{https://huggingface.co/VLABench/pi05-primitive-10task} for finetuned checkpoints, and directly applied them to the simulation experiments.
For GR00T-N1.6, we finetuned \texttt{nvidia/GR00T-N1.6-3B}\footnote{https://huggingface.co/nvidia/GR00T-N1.6-3B} with the training scripts on the official repository\footnote{https://github.com/NVIDIA/Isaac-GR00T/tree/n1.6-release}, using a dataset: \texttt{VLABench/vlabench\_primitive\_ft\_lerobot}\footnote{https://huggingface.co/datasets/VLABench/vlabench\_primitive\_ft\_lerobot}.
We finetuned this model for 100,000 training steps with global batch size of 640, using 8 H100 GPUs.
We also finetuned \texttt{lerobot/pi05\_base}\footnote{https://huggingface.co/lerobot/pi05\_base} for the real-world experiments. 

\begin{figure}[h]
  \centering
  \begin{subfigure}[b]{0.4\textwidth}
    \centering
    \includegraphics[width=\linewidth]{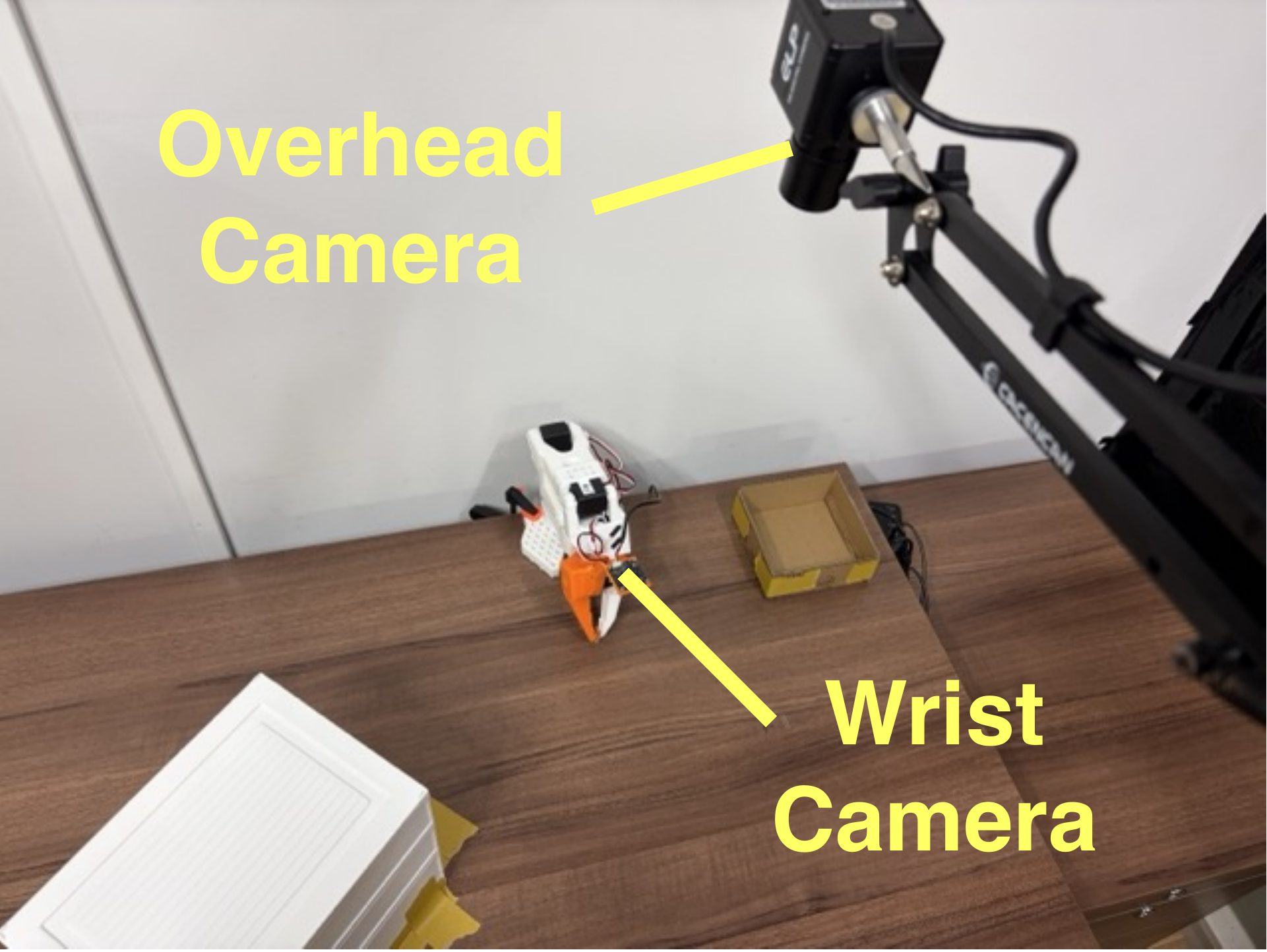}
    \label{figure: real-world-camera}
  \end{subfigure}
  \begin{subfigure}[b]{0.4\textwidth}
    \centering
    \includegraphics[width=\linewidth]{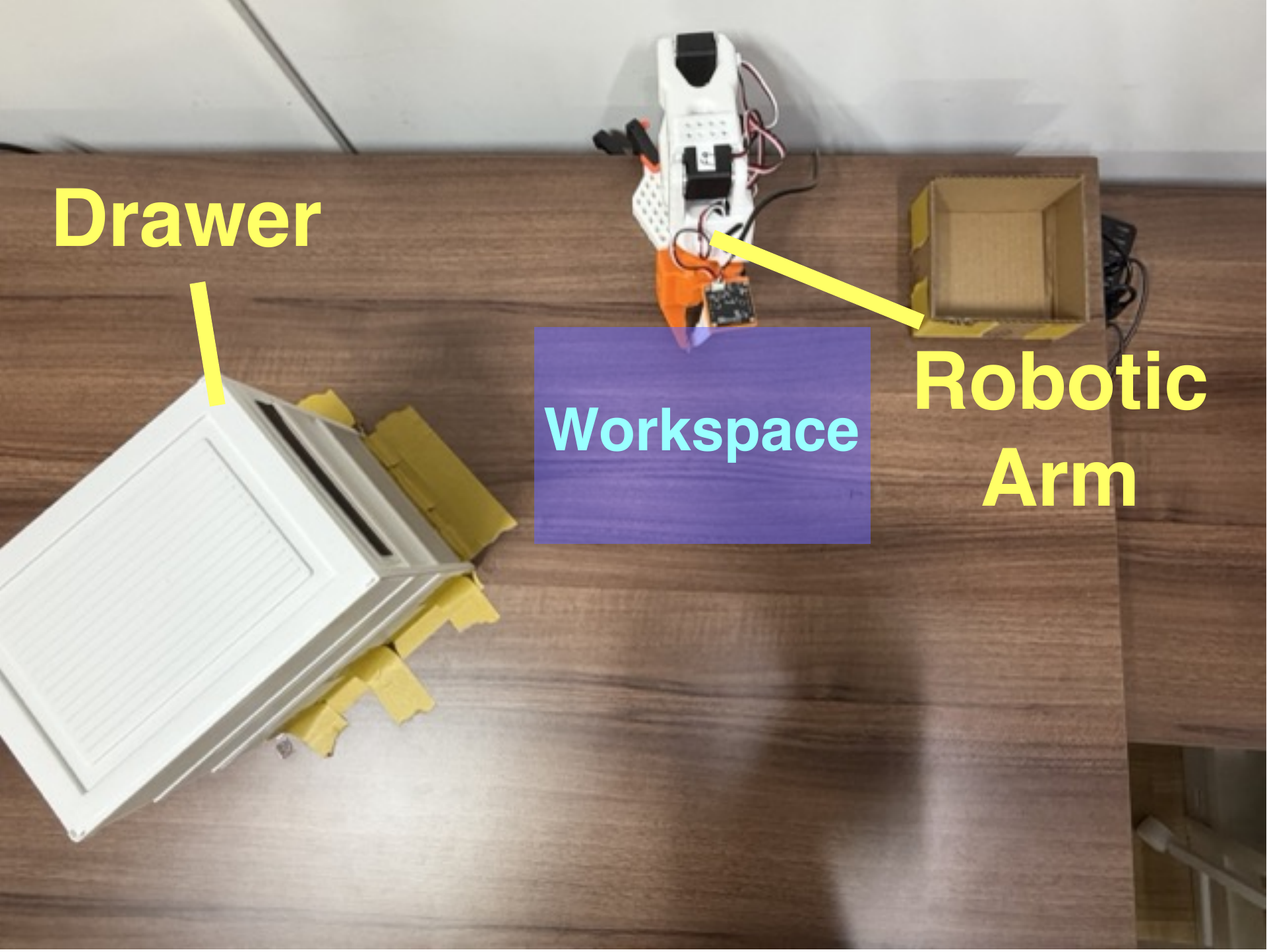}
    \label{figure: real-world-workspace}
  \end{subfigure}
  \caption{\textbf{Real-world Experimental Environment.} We use 6-DoF SO-101 robotic arm with an overhead and wrist-mounted camera.}
  \label{figure: real-world-env}
\end{figure}

\begin{table*}[t]
    \centering
    \caption{
        \textbf{Default hyperparameters and model-specific implementation
        details of ActionCache.}
        Unless otherwise specified, these settings are used throughout
        our experiments.
    }
    \label{tab: default_hyperparameters}
    \small
    \setlength{\tabcolsep}{5pt}
    \renewcommand{\arraystretch}{1.15}

    \begin{tabularx}{\textwidth}{
        @{}
        >{\raggedright\arraybackslash}p{3.2cm}
        c
        >{\raggedright\arraybackslash}p{2.8cm}
        >{\raggedright\arraybackslash}p{3.3cm}
        X
        @{}
    }
        \toprule
        Quantity
        & Symbol
        & $\pi_{0.5}$
        & GR00T-N1.6
        & Description \\
        \midrule

        Cache key source
        & --
        & VLM output
        & VLM output $+$ robot-state
        & Features used as input to the random projection. \\

        Cache key source dimension
        & $D$
        & $1{,}982{,}464$
        & $593{,}408$
        & Dimension of the concatenated feature vector before random
          projection. (model specific) \\

        Cache-key dimension
        & $d$
        & 500
        & 500
        & Dimension of the projected cache key. \\

        Nonzero density
        & $p$
        & 0.01
        & 0.01
        & Fraction of nonzero entries in the sparse ternary random
          projection matrix. \\

        Default hit threshold
        & $T_{\mathrm{hit}}$
        & 0.85
        & 0.65
        & Minimum cosine similarity required to accept a retrieved
          cache entry. \\

        Default cache capacity
        & $C$
        & 3,000
        & 3,000
        & Maximum number of cache entries. \\

        Replacement policy
        & --
        & LRU
        & LRU
        & Default cache replacement policy. \\

        Full-generation NFE
        & $N$
        & 10
        & 4
        & The official number of function evaluations. \\

        Action chunk horizon
        & $H$
        & 10
        & 50
        & Number of actions generated in each action chunk. \\

        Action execution horizon
        & --
        & 10
        & 10
        & Number of actions executed before replanning. \\

        \bottomrule
    \end{tabularx}
\end{table*}

\subsection{Hyperparameter Settings}
\Cref{tab: default_hyperparameters} summarizes the default hyperparameters for ActionCache specified in \Cref{subsec: eval-setting}, and some other specific values used in experiments.
Since VLM output embeddings of GR00T-N1.6 does not contain robot-state features, we utilized concatenation of VLM output and robot-state features as key source.

\subsection{Simulation Evaluation}
\textbf{Random Seed Configuration and Statistical Aggregation.}
For simulation experiments, we controlled two types of seeds: a task-setting seed and an action-noise sampling seed.
Across all experiments on VLABench and LIBERO, the task-setting seed and episode ID uniquely determine the task configuration—including the target object, initial object poses, and instruction—ensuring that no evaluation episode exactly matches those used for cache prefill.
We repeat each evaluation using four noise-sampling seeds, with 200 episodes per seed, and report the mean and standard deviation of the four resulting success rates in \Cref{tab:comp-sr-latency}.
Since the success rate is relatively insensitive to the noise-sampling seed and our goal is to obtain a coarse characterization of the latency–success-rate trade-off, we evaluate EfficientVLA and Falcon using a single noise-sampling seed, corresponding to 200 episodes in total.

\textbf{Evaluation Task on VLABench.}
For our experiments on VLABench, we used 10 primitive tasks of varying difficulty from VLABench, which are released as an official unified dataset on Hugging Face.
All the task names are as follows; \texttt{add\_condiment}, \texttt{insert\_flower}, \texttt{select\_book}, \texttt{select\_chemistry\_tube} (denoted as \texttt{select\_tube} in \Cref{subsec: eval-sr-latency}), \texttt{select\_drink}, \texttt{select\_fruit}, \texttt{select\_mahjong}, \texttt{select\_painting}, \texttt{select\_poker}, and \texttt{select\_toy}.

\subsection{Real-world Training and Evaluation}
\textbf{Real-world Environment.}
As depicted in \Cref{figure: real-world-env}, we use an overhead camera and a wrist-mounted camera with a resolution of 480$\times$640, resized to 224$\times$224 before inputting them to the model.

\textbf{Detailed Task Description.}
We defined three tasks; \texttt{pick\_sausage}, \texttt{pick\_and\_close} and \texttt{push\_button}.
In \texttt{pick\_sausage}, the robot picks up a toy sausage from workspace containing multiple toy foods, and place it on a fixed pan.
There placed two distractor objects to assess the robustness and spatial understanding of ActionCache in perturbations of object layout.
In \texttt{pick\_and\_close}, the robot picks up a white cube, put it inside a drawer, and close the drawer.
This task provides a multi-step and long-horizon evaluation.
In \texttt{push\_button}, three colored buttons (red, green, and blue) are placed in a randomized spatial arrangement, and the robot is instructed to press the button of the specified color.
This task evaluates the model's instruction following and visual/linguistic grounding ability.

\textbf{Training Dataset Construction and Evaluation Settings.}
To construct fine-tuning datasets for each tasks, we record $100$ successful episodes and 50 additional episodes consisting of recovery trajectories from near-failure states.

For \texttt{pick\_and\_close} task, we predetermined five fixed locations to place a white cube and record 20 successful demonstrations per location to ensure comprehensive spatial coverage.
For each location, we add rotational perturbations to the cube; $0\tcdegree$, $\pm30\tcdegree$ and $\pm45\tcdegree$.
In each evaluation episode, one of the five locations is uniformly selected and the same rotational perturbations are applied.
A trial is deemed successful if the robot put the cube into the drawer and close it.

For \texttt{pick\_sausage} task, we randomly choose the location of the sausage and add rotational perturbations; $0\tcdegree$, $45\tcdegree$, $90\tcdegree$, $135\tcdegree$ and $180\tcdegree$.
The location of the other distractor objects are also randomly selected.
The task setting on evaluation is largely consistent with that used for fine-tuning dataset collection; however, object positions are randomized, and none of the evaluated configurations exactly matches a configuration included in the fine-tuning dataset.
A trial is deemed successful if the robot put the sausage on the pan.

For \texttt{push\_button} task, three colored buttons (red, green, and blue) are placed at three predefined positions in front of the robotic arm.
At the beginning of each trial, the assignment of colors to these positions is uniformly randomized, and the target button color specified in the language instruction is also sampled uniformly at random.
A trial is deemed successful if the robot presses the button whose color matches the instruction.

Along with the success rate, we measure the inference wall-clock time to accurately evaluate the latency-performance trade-off. 
We implemented our evaluation code with PyTorch.

\textbf{Training Budget.}
For each task, we fine-tuned $\pi_{0.5}$ for $5{,}000$ steps with a global batch size of $128$.
We used LeRobot framework with $4$ H100 ($96$GB memory) GPUs for fine-tuning.

\textbf{Evaluation Settings.}
Before success rate measurement, we prefill the cache with $T_\textrm{hit}=1$.
Once the cache becomes full, we seamlessly move on to the final evaluation with $T_\textrm{hit}$ value predetermined for each tasks.
Specifically, we set $T_\textrm{hit}=0.875$ for the \texttt{pick\_sausage} task to maintain base model's success rate, and $T_\textrm{hit}=0.85$ for the others.
Since the \texttt{push\_button} task is relatively short-horizon, we set the cache size to $300$ instead of $1{,}000$ for the other tasks.
We set both the action chunk and execution horizon to $50$, since it is the default value for the base model (\texttt{lerobot/pi05\_base}).
Each success rates and latencies reported in Table 2 is aggregated over 50 episodes.

\subsection{VLA-Cache Configuration and evaluation settings}
For evaluation of ActionCache with VLA-Cache in \Cref{subsec: vla-cache}, we use the hyperparameter settings shown in \Cref{tab:vla_cache_hyperparameters}.
The default settings yielded little VLM speedup, likely due to differences in the model and benchmark; we therefore use more aggressive visual token reuse parameters that still preserve the success rate.

\begin{table}[h]
    \centering
    \caption{\textbf{Hyperparameter settings used for VLA-Cache.} Default values from the official VLA-Cache GitHub repository are shown in parentheses.}
    \label{tab:vla_cache_hyperparameters}
    \begin{tabular}{l c}
        \toprule
        Hyperparameter & Value \\
        \midrule
        Threshold of patch-wise cosine similarity
            & 0.996 (0.996) \\
        Token pruning layer locations
            & 2, 6, 9, 11, 13, 15 (2, 6, 9, 11) \\
        Reuse-ratio schedule growth factor
            & 1.00 (0.55) \\
        \bottomrule
    \end{tabular}
\end{table}

%% file: app_hyperparam.tex
\begin{figure}[h]
  \centering
  \includegraphics[width=0.98\textwidth]{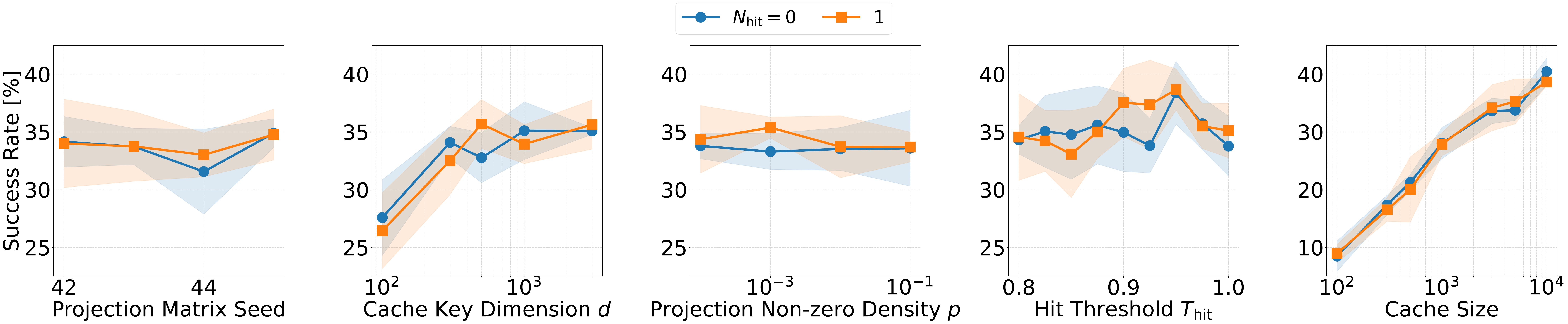}
  \caption{\textbf{Task success rates for hyperparameter changes.} Only one parameter varies and the others are fixed. Cache key dimension $d$ and cache size mainly affect the success rate, while the others do not.}
  \label{figure: sr-vs-hyper}
\end{figure}

\begin{figure}[h]
  \centering
  \includegraphics[width=0.95\textwidth]{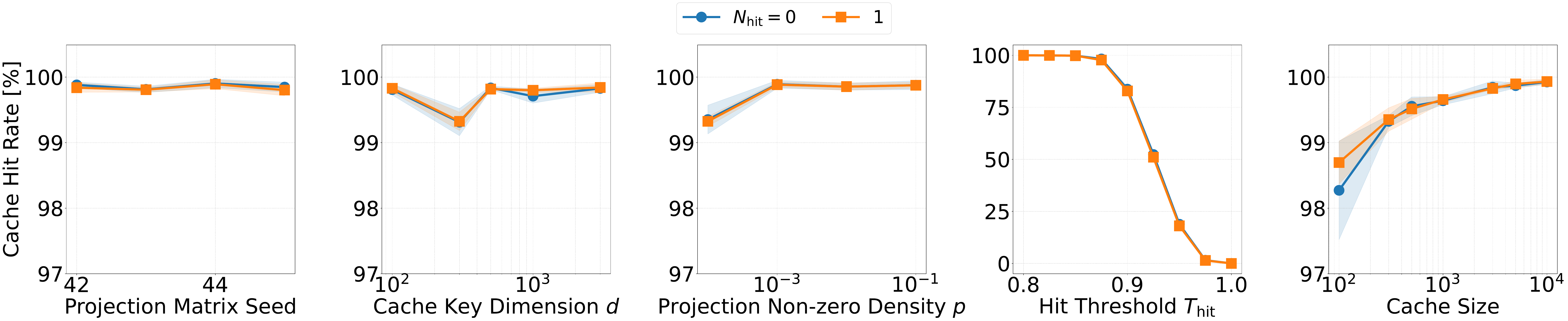}
  \caption{\textbf{Cache hit rates for hyperparameter changes.} Only one parameter varies and the others are fixed. The hit threshold $T_\textrm{hit}$ mainly affects the cache hit rate.}
  \label{figure: hr-vs-hyper}
\end{figure}

\section{Hyperparameter Sensitivity Analysis}\label[appendix]{appsec:hyperparam}

We analyze how hyperparameters affect the task success rates and the cache hit rates of ActionCache.
Specifically, we vary only one hyperparameter among the generation seed of the sparse ternary random matrix, key dimension $d$, non-zero value density of the sparse ternary random matrix $p$, hit threshold $T_\textrm{hit}$ and maximum cache size.
The other hyperparameters are fixed to their default value as specified in \Cref{subsec: eval-setting}.
To focus on the low-NFE regime, we set $N_\textrm{hit} \in \{0,1\}$.
All experiments are conducted on VLABench with $\pi_{0.5}$.

\Cref{figure: sr-vs-hyper} shows the success rates for each hyperparameter variations.
We observe that the overall success rate heavily depends on the cache size.
Notably, cache size of $10,000$ reaches the success rate of the full-step $\pi_{0.5}$ model.
This result suggests that the cache size is the most important factor in our framework.
The cache key dimension affects the success rate only when it is too small, but we found that $d = 300$ is sufficient.
In contrast, the generation seed and non-zero value density of the key projection matrix has little effect on the success rate within the evaluated range.
These results suggest that we can achieve high performance with relatively small values of $d$ and $p$, resulting in negligible computational and memory overhead of the key projection in ActionCache.
The hit threshold $T_\textrm{hit}$ also has little effect in this evaluation setting.
This is likely because, with sufficient cache size ($3,000$ for this time), the quality of a Top-1 candidate becomes high enough on average, making the success rate largely insensitive to the choice of $T_\textrm{hit}$.

\Cref{figure: hr-vs-hyper} shows the cache hit rate for each hyperparameter variations.
We observe that all hyperparameters except for $T_\textrm{hit}$ have little effect on the cache hit rate, which remains close to $100$\% across the evaluated range.
On the other hand, $T_\textrm{hit}$ dominates the cache hit rate; the hit rate monotonically decreases from nearly $100$\% to $0$\% as $T_\textrm{hit}$ increases.
Since the cache hit rate directly affects the average inference latency in this framework, $T_\textrm{hit}$ plays an important role in the latency--success-rate trade-off.

%% file: app_longspan.tex
\begin{figure}[h]
  \centering
  \includegraphics[width=0.6\textwidth]{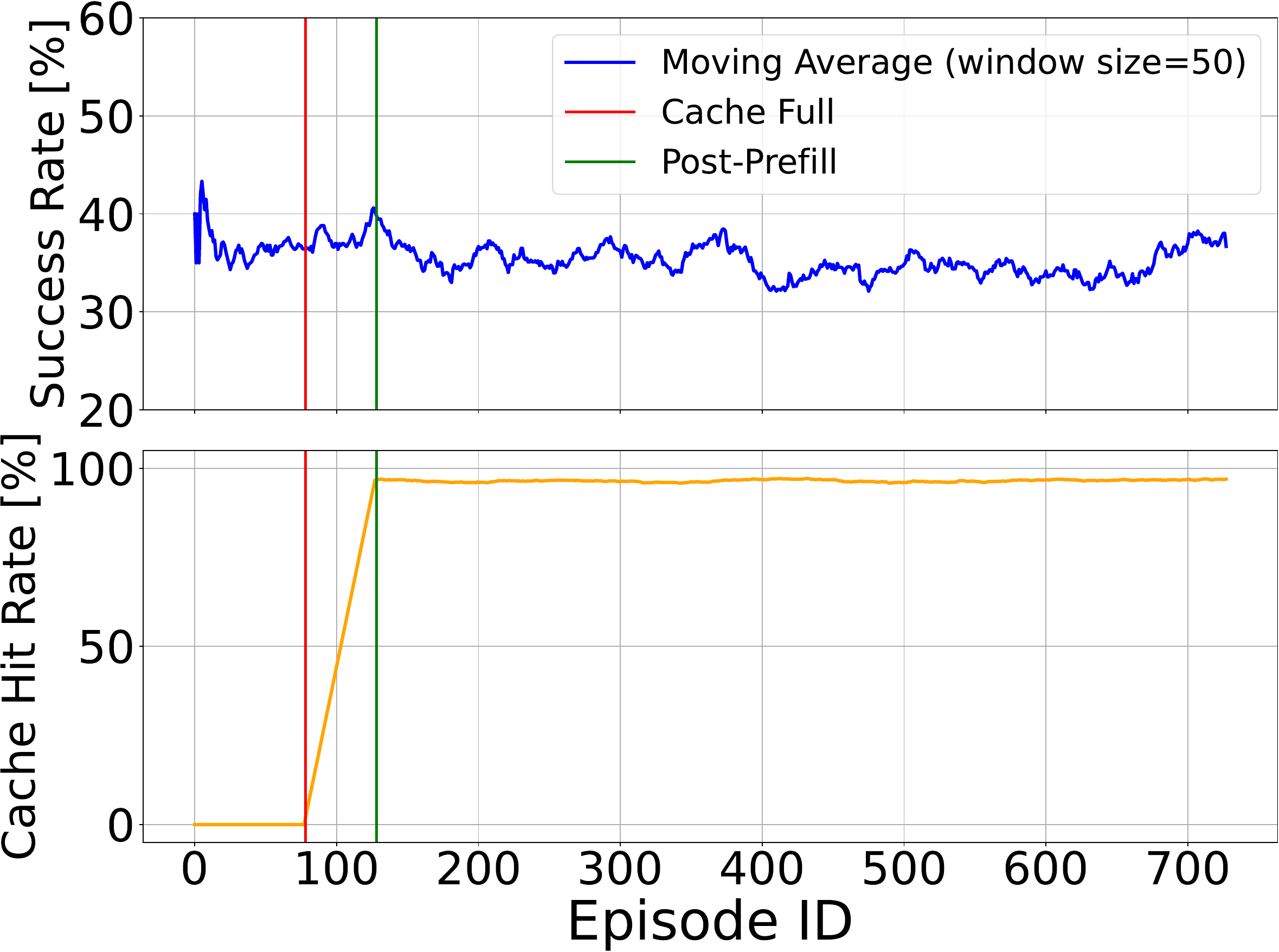}
  \caption{\textbf{Moving average of task success rate and cache hit rate.} The red and green vertical line shows the timings when the cache becomes full and when the moving average no longer contains prefill episodes.}
  \label{figure: sr-hr-mov-avg}
\end{figure}

\section{Long-span Caching Evaluation}\label[appendix]{appsec:longspan}

Once the cache becomes full, ActionCache continuously replaces old or useless actions with new ones.
To evaluate ActionCache behavior in long-span deployment, we conduct a long period experiment on VLABench.
In this experiment, we measure the moving average of task success rate and cache hit rate with a window size of 50 episodes.
We use the hit threshold $T_\textrm{hit}=0.88$ to ensure a certain degree of cache miss and cache replacement, and use $N_\textrm{hit} = 1$.
The other hyperparameters are set to default values as specified in \Cref{subsec: eval-setting}.
We use $\pi_{0.5}$ for this experiment.

\Cref{figure: sr-hr-mov-avg} shows the moving averages of the task success rate and the cache hit rate.
In this figure, even long after the prefill ends, the task success rate and the cache hit rate remain at the same level as immediately after the prefill ends.
This result indicates that ActionCache maintains its efficacy in long span deployment, where cache entries are continuously replaced.

%% file: app_similarity.tex
\section{Key Similarity Analysis in Real-world}
\begin{figure}[t]
    \centering

    \subcaptionbox{
        \texttt{push\_button}
        \label{fig: push-button-flow}
    }{
        \includegraphics[
            width=0.5\linewidth,
            keepaspectratio
        ]{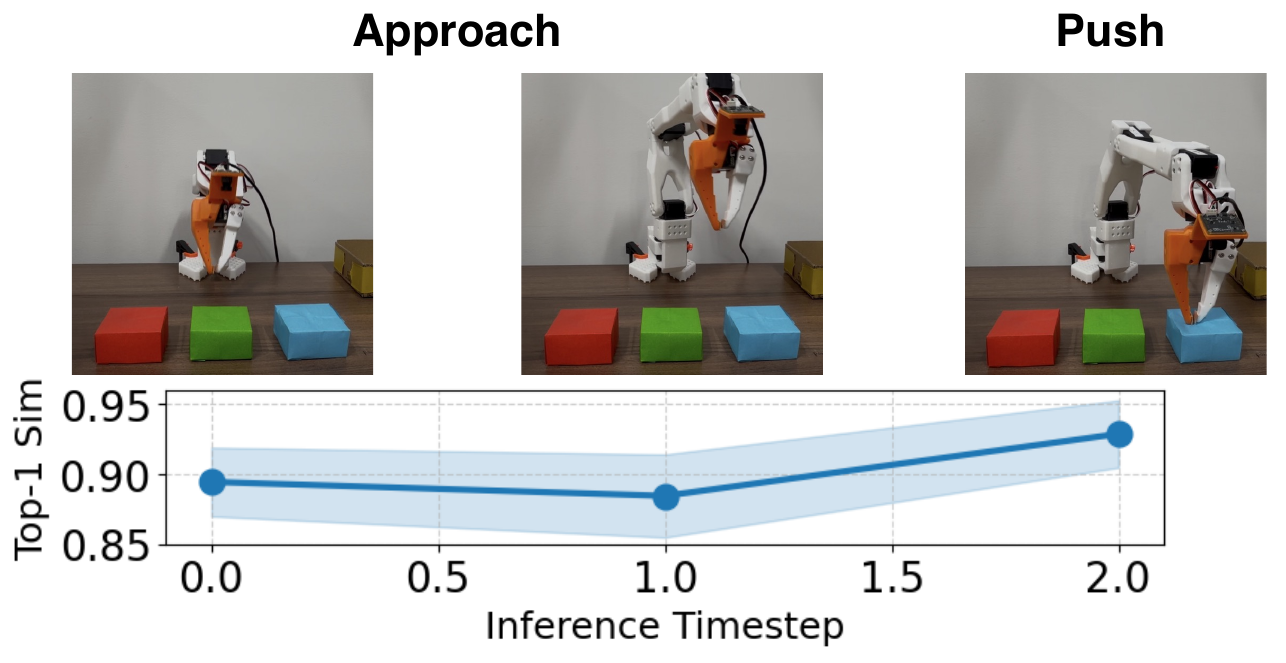}%
    }
    \par\vspace{0.8em}
    \subcaptionbox{
        \texttt{pick\_sausage}
        \label{fig: pick-sausage-flow}
    }{
        \includegraphics[
            width=0.99\linewidth,
            keepaspectratio
        ]{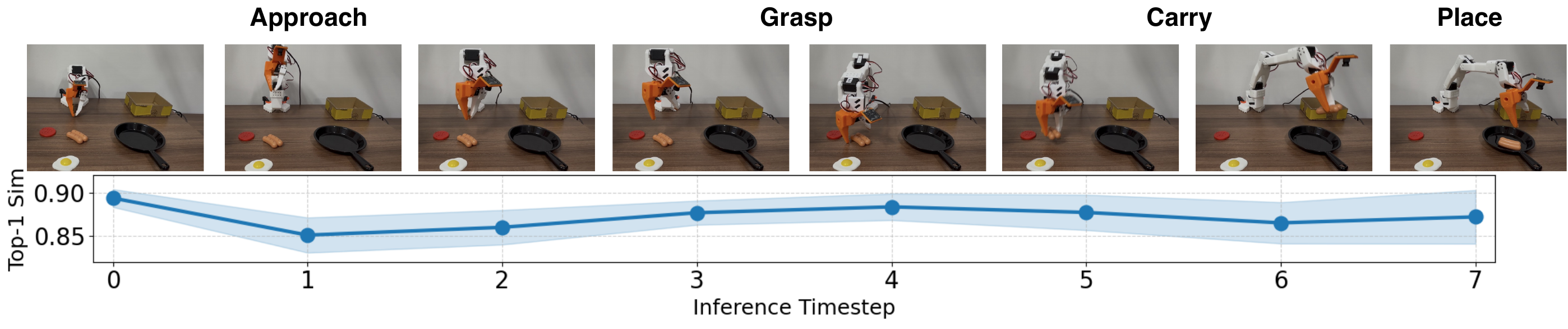}%
    }
    \par\vspace{0.8em}
    \subcaptionbox{
        \texttt{pick\_and\_close}
        \label{fig: pick-and-close-flow}
    }{
        \includegraphics[
            width=0.99\linewidth,
            keepaspectratio
        ]{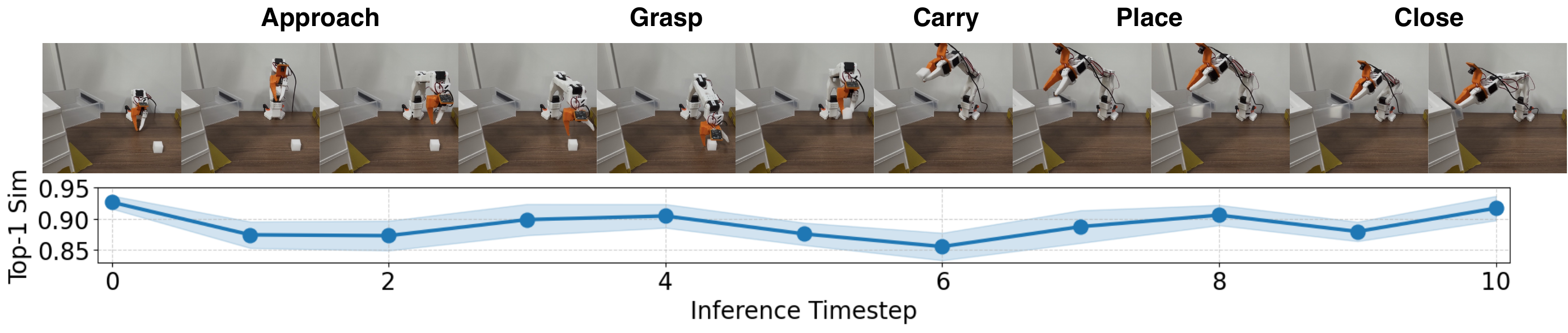}%
    }
    \caption{
        Visualization of Top-1 similarity and task phase
        for the three manipulation tasks.
    }
    \label{fig: manipulation-flows}
\end{figure}

\label[appendix]{appsec:similarity}
We analyze the temporal evolution of the top-1 cache key similarity in real-robot experiments.
Using all data from the real-robot experiments in \Cref{subsec: eval-real-world}, we compute the mean and standard deviation of the top-1 similarity at each inference timestep over successful episodes.
Failed episodes are excluded because they are substantially longer and too few to provide reliable statistics.
\Cref{fig: manipulation-flows} shows the temporal evolution of the top-1 similarity for each task, together with intermediate images corresponding to selected timesteps.

For \texttt{pick\_sausage} and \texttt{pick\_and\_close}, the similarity tends to decrease during phases such as approaching, carrying, and the transition from placing to closing.
These phases involve substantial changes in the camera observation, suggesting that the cache keys are sensitive to changes in visible objects and background appearance.
In contrast, high similarities are observed near the beginning of episodes and during actions shared across many episodes, such as placing, closing, and pushing.
This indicates that ActionCache successfully retains and reuses such common action patterns across episodes.

%% file: app_libero.tex
\begin{table}[t!]
  \centering
  \caption{\textbf{Success rate comparison of $\pi_{0.5}$ and ActionCache on LIBERO.}}
  \label{tab: sr-vs-nfe-libero}
  \small
  \begin{tabular}{lrrrrrrrr}
  \toprule
  \multirow{2}{*}{Method} & \multirow{2}{*}{NFE} & \multicolumn{5}{c}{Success Rate (\%)} & Hit Rate & Latency \\
  \cmidrule(lr){3-7}
  & & Spatial & Object & Goal & Long & Avg. & (\%) & (ms) \\
  \midrule
  \multirow{2}{*}{Base Model}
  & 10 & 97.4 & \textbf{99.0} & 96.6 & 95.4 & \textbf{97.1} & -- & 54.07 \\
  & 1  & \textbf{97.6} & 98.0 & 97.4 & 94.6 & 96.9 & -- & 5.84 \\
  \midrule
  \multirow{3}{*}{\textbf{ActionCache}}
  & 2 & 96.0 & 98.4 & \textbf{97.6} & \textbf{96.0} & 97.0 & 92.57 & 14.25 \\
  & 1 & 96.4 & \textbf{99.0} & 96.2 & 93.6 & 96.3 & 91.88 & 11.31 \\
  & 0 & 89.4 & \textbf{99.0} & 96.4 & 83.4 & 92.1 & 87.73 & 6.57 \\
  \bottomrule
  \end{tabular}
\end{table}

\section{Results for LIBERO benchmark}\label[appendix]{appsec:libero}
LIBERO consists of four task suites (Spatial, Object, Goal and Long) which contain 10 tasks each.
We evaluate $\pi_{0.5}$ on LIBERO benchmark and report its success rate for each task suites at $\mathrm{NFE}=10$ and $1$ for the base model, and $\mathrm{NFE}=0\sim2$ for ActionCache.
We also report cache hit rate and latency of action head.
Action head latency is measured on NVIDIA RTX 5090 GPU with 32GB of memory.
For all task suites, we set the cache size to $10{,}000$ and $T_\textrm{hit}=0.85$.
In this experiment, we leveraged an open weighted fine-tuned checkpoint (\texttt{lerobot/pi05\_libero\_finetuned\_v044}\footnote{https://huggingface.co/lerobot/pi05\_libero\_finetuned\_v044}) from Huggingface, and used LeRobot framework to conduct the experiment.

\textbf{Evaluation Settings.}
The success rates are the mean value on 500 episodes for each task suites and NFE settings.

\textbf{Results.}
As shown in \Cref{tab: sr-vs-nfe-libero}, the base model maintains its full-step success rate at $\textrm{NFE}=1$.
This result shows that easy tasks can be solved with very few denoising steps on flow-based VLA models.
ActionCache also preserves the success rate in the low-NFE regime, especially at $\textrm{NFE=1}$ and $2$.
Together with the results in \Cref{subsec: eval-sr-latency}, this demonstrates that ActionCache is effective whether or not the base model degrades at low-NFE, suggesting its applicability across tasks of varying difficulty.